\documentclass[sn-basic,iicol]{sn-jnl}
\usepackage{graphicx}
\usepackage{amsmath}
\usepackage{amssymb}
\usepackage{mathdots}
\usepackage{relsize}
\usepackage{multirow}
\usepackage{epstopdf}
\usepackage{subfig}
\usepackage{caption}
\usepackage{natbib}
\usepackage{mhchem}
\usepackage{footnote}

\theoremstyle{thmstyleone}%
\theoremstyle{thmstyletwo}%

\theoremstyle{thmstylethree}%

\begin{document}

\title[Article Title]{Machine learning models for determination of weldbead shape parameters for gas metal arc welded T-joints - A comparative study}


\author*[1]{\fnm{Rururaja} \sur{Pradhan}}\email{rururaj.mech@gmail.com}

\author[1]{\fnm{Apurva P.} \sur{Joshi}}\email{apurvajoshi@iitkgp.ac.in}
\equalcont{These authors contributed equally to this work.}

\author[2]{\fnm{Mohammed Rabius} \sur{Sunny}}\email{sunny@aero.iitkgp.ac.in}
\equalcont{These authors contributed equally to this work.}

\author[1]{\fnm{Arunjyoti} \sur{Sarkar}}\email{arun@naval.iitkgp.ac.in}
\equalcont{These authors contributed equally to this work.}

\affil*[1]{\orgdiv{Department of Ocean Engineering and Naval Architecture}, \orgname{Indian Institute of Technology}, \orgaddress{\city{Kharagpur}, \postcode{721302}, \state{W.B}, \country{India}}}

\affil[2]{\orgdiv{Department of Aerospace Engineering},\orgname{Indian Institute of Technology}, \orgaddress{\city{Kharagpur}, \postcode{721302}, \state{W.B}, \country{India}}}


\abstract{The shape of a weld bead is critical in assessing the quality of the welded joint. In particular, this has a major impact in the accuracy of the results obtained from a numerical analysis. This study focuses on the statistical design techniques and the artificial neural networks, to predict the weld bead shape parameters of shielded Gas Metal Arc Welded (GMAW) fillet joints. Extensive testing was carried out on low carbon mild steel plates of thicknesses ranging from 3mm to 10mm. Welding voltage, welding current, and moving heat source speed were considered as the welding parameters. Three types of multiple linear regression models (MLR) were created to establish an empirical equation for defining GMAW bead shape parameters considering interactive and higher order terms. Additionally, artificial neural network (ANN) models were created based on similar scheme, and the relevance of specific features was investigated using SHapley Additive exPlanations (SHAP). The results reveal that MLR-based approach performs better than the ANN based models in terms of predictability and error assessment. This study shows the usefulness of the predictive tools to aid numerical analysis of welding.}




\maketitle

\section{Introduction}\label{sec1}

Gas metal arc welding (GMAW) is a commonly practiced metal joining process, used predominantly in marine construction industries. High temperature beyond a melting point is generated when an electric arc is established between the metal electrode and base metal. No flux is used as the shielding gas (Ar, He, Co, etc.) is used to protect the weld pool from the contaminants in the atmosphere. The shielding gas technique is widely used in heavy industries due to its ease of automation to fabricate large stiffened panels (\cite{Gadagi2019}) where stiffeners are fillet-jointed to the plate. Weld-induced distortion and residual stresses are the main concerns for  such panels. As a result, numerical simulations are frequently used to evaluate the likely behavior during the manufacture of stiffened panels, and subsequently to select a welding sequence in order to minimize the weld-induced distortion and residual stresses. Accurate estimation of temperature distribution over the plate with respect to time is critical in assessing the weld induced distortion and residual stresses using numerical analysis. Size of the weld bead in the computational domain plays an important role for the accuracy of  numerical model and subsequently on the quality of fabricated product.

In this regard, various researches have attempted to establish input-output correlations of welding parameters and shape of the weld bead for different types of butt welding process through analysis and processing of experimental data statistically (\cite{Apps1963},\cite{Becker1978},\cite{McGlone1982},\cite{Pandey1989},\cite{Senthil1986}). Methods like, linear regression, multiple linear regression, non-linear regression, response surface methodology, Taguchi method etc are commonly used in practice(\cite{Wuest2016},\cite{praga},\cite{deb2012optimization}). \cite{Tarng1998} used Taguchi method to establish an optimized relation for butt weld bead geometry in GTAW. Following similar approach, \cite{Kim2003a} created a model of process parameters of back-bead width in GMAW in mild steel. An extensive comparison between MLR and neural network methods was also carried out.
\cite{Ganjigatti2007} explained a full factorial design to correlate the metal inert gas parameters on "bead-on-plate" type weld of mild steel plates. Linear as well as nonlinear regression analysis were developed and their performances were compared.

Artificial neural networks have recently evolved and have proven to be useful in solving a variety of engineering challenges.\cite{Nagesh2010}reported that neural networks are effective tools for predicting butt weld bead geometry in TIG welding process.\cite{Sathiya2012}also developed an ANN model to establish the relationship between the welding process parameters and the butt weld bead geometry in laser welding. Similar approach was followed by \cite{campbell2012artificial}, \cite{Sarkar2016} on GMAW using alternating shielding gasses.

Based on the literature review, it is observed that various works related to bead geometry modelling for butt welding using GMAW are reported, whereas similar studies on fillet welds are scarce in the literature. In general, any large steel structure construction could involve large length of fillet welded joints, and hence  it is essential that a systematic study is undertaken to evaluate the fillet weld bead shape with respect to the major input weld parameters. This gap has been addressed in  the current article, and a comparative analysis is carried out between various MLR models and ANN models for fillet weld bead geometry. The proposed models have exhibited excellent performance in emulating the fillet weld bead shape based on the major input weld parameters.

\section{Approach to the problem}\label{sec2}

\begin{figure}[h]
	\centering
	\includegraphics[width=0.9\linewidth]{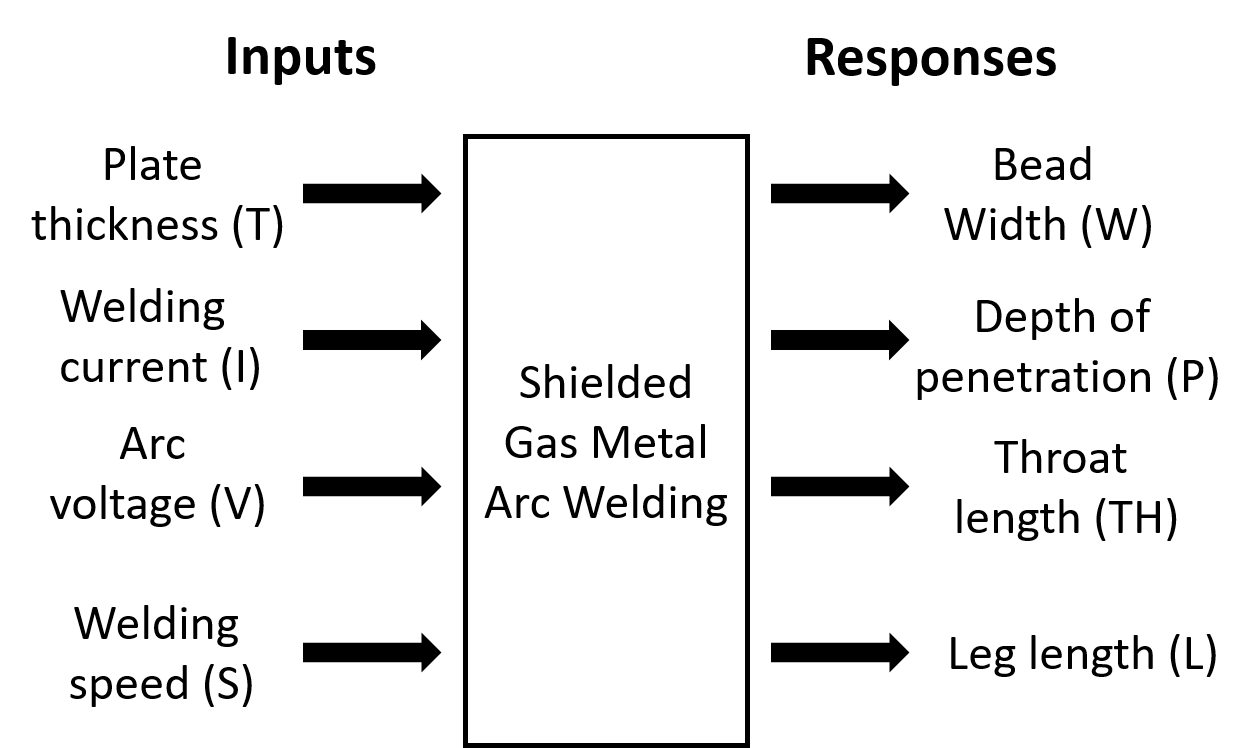}
	\caption{Input-response flow diagram of GMAW}
	\label{fig:picture1}
\end{figure}

The objective of the current work is to predict the input-output relationships of a shielded GMAW process as shown in Figure \ref{fig:picture1}. The welding process parameters and response variables (bead size parameters) are designated as inputs and outputs for the process. Two approaches are considered for this work, namely (a) MLR, where each response will have a single set of relation with the input variable and (b) ANN, using the same set of data.
\begin{figure}[h]
	\centering
	\includegraphics[width=0.9\linewidth]{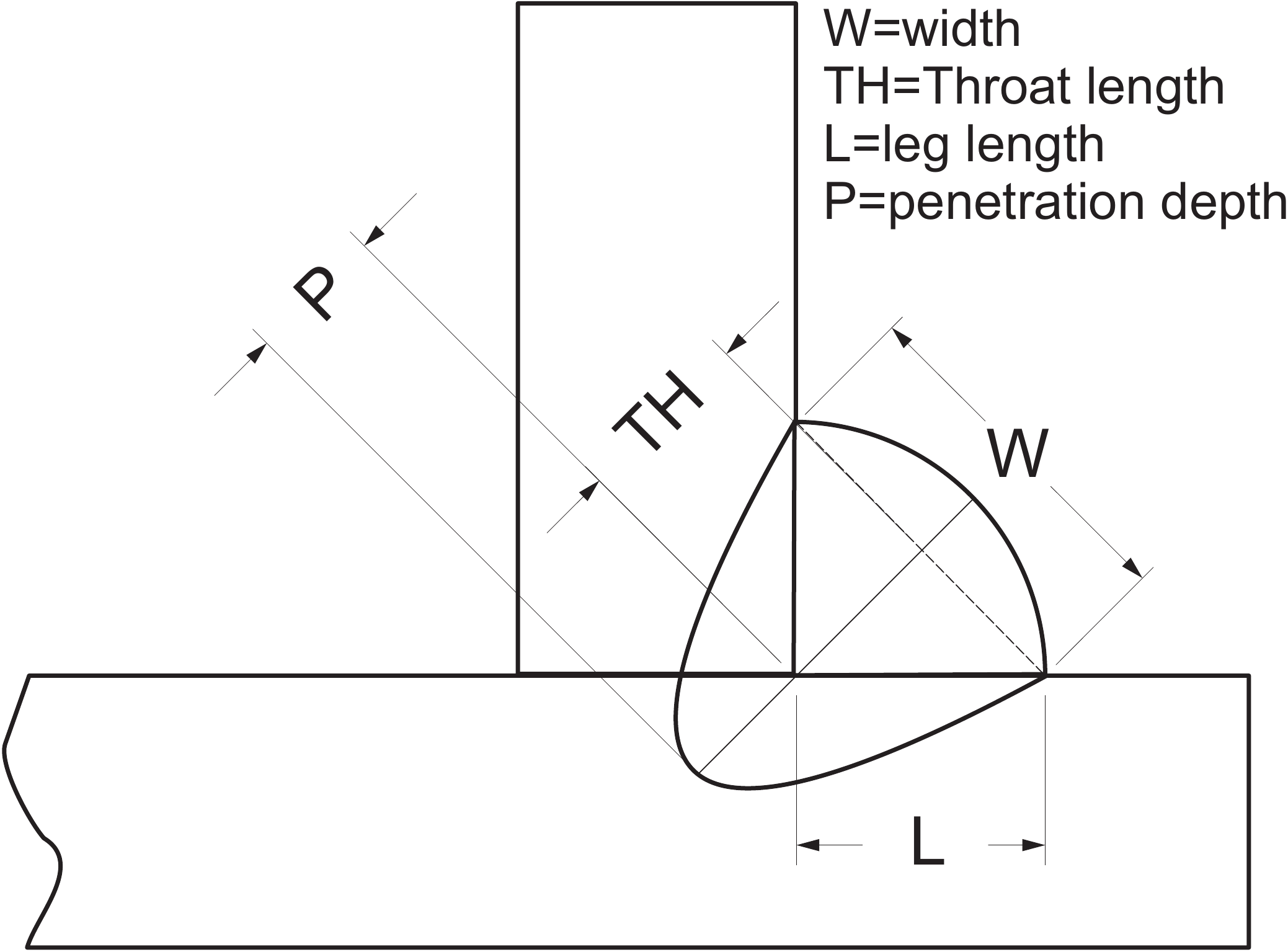}
	\caption{Weld bead shape parameters}
	\label{fig:picture2}
\end{figure}
As shown in Figure \ref{fig:picture1}, the chosen input parameters for welding in this work are: plate thickness (T), welding current (I), arc voltage (V) and welding speed (S). Similarly, the output responses are: bead width (W), depth of penetration (P), throat length (TH) and leg length (L). In this case it is assumed that the leg height and leg length to be same. The schematic of the bead shape parameters is shown in Figure \ref{fig:picture2}. It is assumed that the weld pool produced by the welding arc is the principal source of heat.


\section{Generation of data for predictive model}\label{sec3}
\begin{figure}[h]
	\centering
	\includegraphics[width=0.5\linewidth]{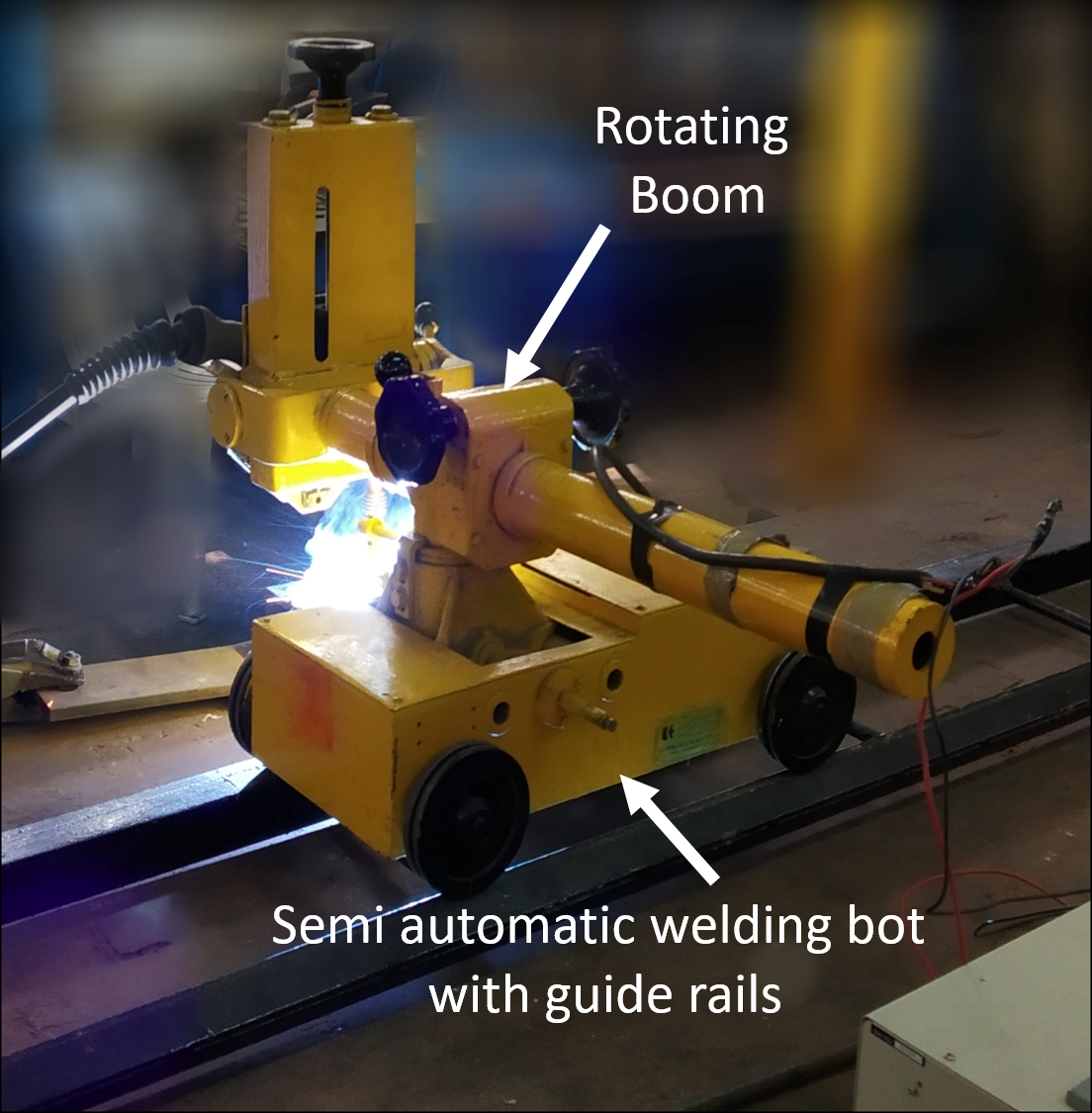}
	\caption{The welding set up }
	\label{fig:picture3}
\end{figure}

In the present study, low carbon mild steel (0.16\% C) plates of thickness 3mm, 4mm, 5mm, 6mm, 8mm and 10 mm were fillet welded using a semiautomatic GMAW machine. Figure \ref{fig:picture3} shows the semi-automatic setup of the welding station. Welding was carried out using constant voltage electrode positive polarity with flux cored wire of diameter 1.2 mm. 
The diameter of the feed wire and wire feed rate were kept constant throughout the experiments. The shielding gas used in GMAW was 100\% \ce{CO2}. The gas flow rate was 1.2 lpm and the gas preheat temperature was 80°C. In the fillet welding experiments, the parameters recorded are current (I) flowing through the circuit, voltage (V) of the arc and torch travel speed (S). These data are listed in Table \ref{table1} and Table \ref{table2}.
\begin{figure}[h]
	\centering
	\subfloat[]{%
		\includegraphics[width=0.7\linewidth, height=0.3\textwidth]{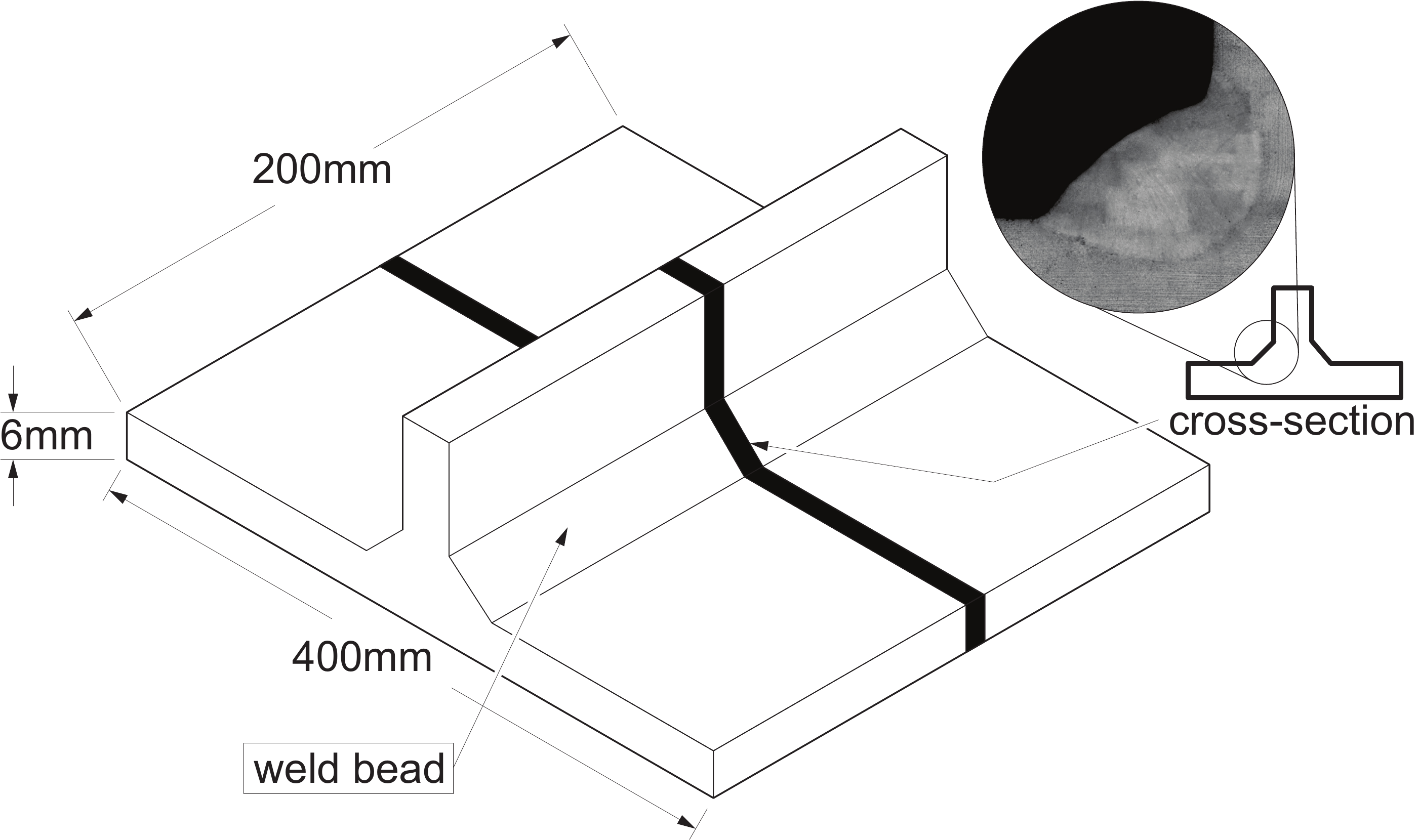}
		\label{fig:picture4a}}
	
	\subfloat[]{
		\includegraphics[width=0.4\linewidth, height=0.25\textwidth]{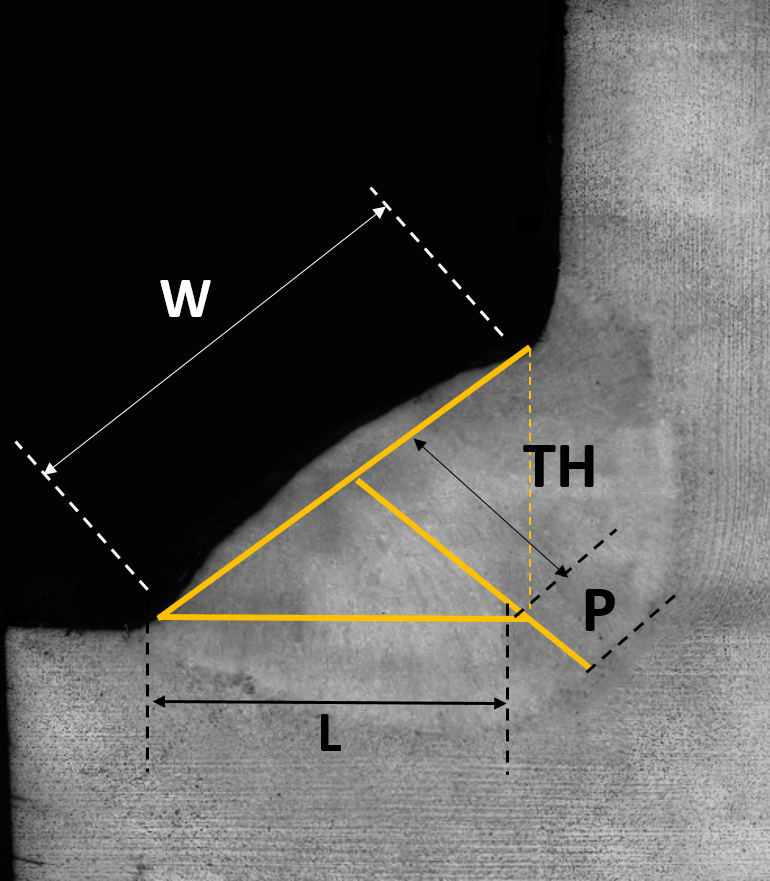}
		\label{fig:picture4b}}
	\caption{(a)Schematic diagram for sample preparation area in fillet welding (b) Weld deposit showing response parameters }
	\label{fig:picture4}
\end{figure} 
Weld bead samples for macroscopic inspection are cut from the middle of the plate (Figure \ref{fig:picture4a}) to nullify the end effects. The transverse surface of the cut samples were then mirror finished and subsequently etched with 2\% Nital solution for macroscopic observations as shown in Figure \ref{fig:picture4b}.

\begin{table*}
	\centering
	\caption{Welding process parameters and response variable data set for training}
	\label{table1}
	\resizebox{\textwidth}{!}{%
		\begin{tabular}{ccccccccc}
			\hline
			& \multicolumn{4}{c}{Input} & \multicolumn{4}{c}{Output} \\ \cline{2-9} 
			Number & \begin{tabular}[c]{@{}c@{}}Thickness   of plate \\ (mm)\end{tabular} & \begin{tabular}[c]{@{}c@{}}Welding   current\\  (A)\end{tabular} & \begin{tabular}[c]{@{}c@{}}Arc   voltage\\    (V)\end{tabular} & \begin{tabular}[c]{@{}c@{}}Welding   speed\\    (mm/s)\end{tabular} & \begin{tabular}[c]{@{}c@{}}Width\\    (mm)\end{tabular} & \begin{tabular}[c]{@{}c@{}}Depth of   penetration\\    (mm)\end{tabular} & \begin{tabular}[c]{@{}c@{}}Throat   length\\    (mm)\end{tabular} & \begin{tabular}[c]{@{}c@{}}Leg   length\\    (mm)\end{tabular} \\ \cline{2-9} 
			& (T) & (I) & (V) & (S) & (W) & (P) & (TH) & (L) \\ \hline
			1 & 10 & 310 & 28 & 9.25 & 4.7 & 1.8 & 4.2 & 4.9 \\
			2 & 3 & 175 & 25 & 7 & 5 & 1 & 4 & 4.4 \\
			3 & 6 & 175 & 18 & 5 & 5.6 & 1.2 & 4.6 & 4.3 \\
			4 & 4 & 125 & 30 & 4 & 4.1 & 1.6 & 5 & 4.5 \\
			5 & 6 & 275 & 30 & 8 & 5 & 1.8 & 4.8 & 5 \\
			6 & 8 & 200 & 34 & 5 & 4.4 & 2.1 & 4.7 & 5.2 \\
			7 & 3 & 250 & 28 & 8 & 5.3 & 1.5 & 4.6 & 4.7 \\
			8 & 4 & 165 & 27 & 5.25 & 4.8 & 1.5 & 4.7 & 4.5 \\
			9 & 8 & 250 & 32 & 6.75 & 4.9 & 1.9 & 4.5 & 5 \\
			10 & 3 & 220 & 30 & 4 & 5.2 & 1.8 & 5.4 & 5.2 \\
			11 & 3 & 230 & 30 & 8 & 5.2 & 1.6 & 4.6 & 4.6 \\
			12 & 5 & 300 & 22 & 6 & 7 & 2 & 5.2 & 5.4 \\
			13 & 3 & 225 & 30 & 8 & 5.2 & 1.5 & 4.5 & 4.6 \\
			14 & 8 & 200 & 26 & 6 & 5.1 & 1.5 & 4.2 & 4.7 \\
			15 & 4 & 270 & 20 & 4 & 7 & 1.8 & 5.8 & 5.2 \\
			16 & 8 & 275 & 22 & 9 & 4.8 & 1.3 & 3.8 & 5 \\
			17 & 6 & 220 & 24 & 9.25 & 4.5 & 1.1 & 3.5 & 4.5 \\
			18 & 8 & 190 & 32 & 8 & 5.2 & 1.5 & 4 & 4.2 \\
			19 & 5 & 190 & 30 & 7 & 5 & 1.5 & 4.2 & 4.5 \\
			20 & 5 & 150 & 20 & 5.5 & 4.8 & 1 & 4 & 4.2 \\
			21 & 10 & 175 & 32 & 3.25 & 4 & 2.2 & 5 & 5.3 \\
			22 & 10 & 225 & 26 & 4 & 5.7 & 2 & 4.9 & 5.2 \\
			23 & 4 & 200 & 34 & 6 & 4.8 & 1.8 & 4.8 & 4.9 \\
			24 & 5 & 175 & 36 & 6 & 4.7 & 1.9 & 4.8 & 4.7 \\
			25 & 5 & 200 & 34 & 5 & 4.5 & 2.1 & 5 & 5.2 \\
			26 & 5 & 310 & 28 & 9.25 & 4.9 & 1.8 & 5 & 4.8 \\
			27 & 8 & 175 & 32 & 5 & 4.6 & 1.9 & 4.6 & 4.9 \\
			28 & 4 & 175 & 28 & 5 & 4.9 & 1.6 & 4.8 & 4.6 \\
			29 & 5 & 230 & 26 & 6 & 5.5 & 1.8 & 4.7 & 5 \\
			30 & 10 & 250 & 32 & 6.75 & 4.8 & 2 & 4.3 & 5 \\
			31 & 6 & 280 & 25 & 8 & 5.4 & 1.7 & 4.5 & 5 \\
			32 & 6 & 175 & 32 & 5 & 4.6 & 1.9 & 4.6 & 4.9 \\
			33 & 5 & 310 & 34 & 9.25 & 4.7 & 1.9 & 5.6 & 4.5 \\
			34 & 8 & 310 & 28 & 9.25 & 4.7 & 1.8 & 4.6 & 4.9 \\
			35 & 3 & 150 & 26 & 5 & 4.7 & 1.3 & 4.8 & 4.3 \\
			36 & 4 & 220 & 22 & 5 & 6.2 & 1.7 & 5.2 & 4.8 \\
			37 & 3 & 200 & 32 & 7 & 5.2 & 1.6 & 4.7 & 4.5 \\
			38 & 8 & 175 & 36 & 6 & 4.6 & 2 & 4.6 & 4.7 \\
			39 & 6 & 310 & 34 & 9.25 & 4.6 & 2 & 5.4 & 4.6 \\
			40 & 6 & 310 & 28 & 9.25 & 4.8 & 1.9 & 4.7 & 4.9 \\
			41 & 5 & 275 & 30 & 8 & 5.2 & 1.9 & 4.9 & 4.9 \\
			42 & 8 & 280 & 35 & 5 & 4.4 & 2.2 & 4.7 & 5.8 \\
			43 & 3 & 250 & 20 & 8 & 5.5 & 1.1 & 4.1 & 5 \\
			44 & 4 & 150 & 32 & 7 & 5 & 1.4 & 4.2 & 4.2 \\
			45 & 4 & 141 & 30 & 8 & 5.2 & 1 & 3.7 & 4 \\
			46 & 10 & 300 & 20 & 7 & 6.4 & 1.8 & 4.1 & 5 \\
			47 & 4 & 180 & 30 & 6 & 5 & 1.6 & 4.6 & 4.6 \\
			48 & 3 & 180 & 25 & 6 & 5.1 & 1.3 & 4.5 & 4.5 \\
			49 & 6 & 150 & 22 & 6 & 4.6 & 1 & 3.8 & 4.3 \\
			50 & 4 & 150 & 26 & 5 & 4.7 & 1.4 & 4.7 & 4.4 \\
			51 & 5 & 200 & 26 & 6 & 5.2 & 1.4 & 4.5 & 4.6 \\
			52 & 8 & 160 & 25 & 6 & 4.7 & 1.2 & 4 & 4.4 \\
			53 & 6 & 200 & 26 & 6 & 5.2 & 1.5 & 4.4 & 4.7 \\ \hline
		\end{tabular}%
	}

\end{table*}
\begin{table*}[]
	\centering
	\caption{Welding process parameters and response variable data set for testing}
	\label{table2}
	\resizebox{\textwidth}{!}{%
		\begin{tabular}{ccccccccc}
			\hline
			& \multicolumn{4}{c}{Input} & \multicolumn{4}{c}{Output} \\ \hline
			Number & \begin{tabular}[c]{@{}c@{}}Thickness  of plate \\ (mm)\end{tabular} & \begin{tabular}[c]{@{}c@{}}Welding   current\\    (A)\end{tabular} & \begin{tabular}[c]{@{}c@{}}Arc  voltage\\    (V)\end{tabular} & \begin{tabular}[c]{@{}c@{}}Welding   speed \\    (mm/s)\end{tabular} & \begin{tabular}[c]{@{}c@{}}Width\\    (mm)\end{tabular} & \begin{tabular}[c]{@{}c@{}}Depth of   penetration\\    (mm)\end{tabular} & \begin{tabular}[c]{@{}c@{}}Throat   length\\    (mm)\end{tabular} & \begin{tabular}[c]{@{}c@{}}Leg   length\\    (mm)\end{tabular} \\ \cline{2-9} 
			& (T) & (I) & (V) & (S) & (W) & (P) & (TH) & (L) \\ \hline
			1 & 10 & 280 & 28 & 6 & 5.5 & 2.2 & 4.5 & 5.2 \\
			2 & 10 & 200 & 34 & 5 & 4.3 & 2.1 & 4.5 & 5.2 \\
			3 & 3 & 125 & 30 & 4 & 4.2 & 1.6 & 5.2 & 4.5 \\
			4 & 6 & 300 & 30 & 7 & 5.2 & 2.1 & 5 & 5.3 \\
			5 & 3 & 275 & 26 & 9.25 & 5 & 1.4 & 4.5 & 4.8 \\
			6 & 6 & 250 & 32 & 6.75 & 5 & 2 & 4.7 & 5 \\
			7 & 4 & 310 & 28 & 9.25 & 4.9 & 1.7 & 5.2 & 4.8 \\
			8 & 4 & 250 & 24 & 6.75 & 5.78 & 1.7 & 4.7 & 5 \\
			9 & 5 & 250 & 32 & 6.75 & 5 & 2 & 4.8 & 5 \\
			10 & 10 & 310 & 34 & 9.25 & 4.4 & 2 & 4.8 & 4.6 \\ \hline
		\end{tabular}%
	}
\end{table*}


\section{Data preprocessing for MLR and ANN}\label{sec4}

The experimental data are tabulated to examine the effects of input variables on the bead geometry parameters of the GMAW process. Pre-processing of data is imperative in the case of any predictive data model like MLR and ANN. This provides utility functions and transformer classes to change raw vectors into more usable predictors. For that purpose, an inhouse Python script is developed, and a well-known machine learning package, “scikit-learn,” is employed for the predictive data analysis (for MLR and preprocessing of ANN). This module contains extensive machine learning algorithms integrated with Python (\cite{Pedregosa2011}).

To avoid the risk of data leakage, the entire data set is randomly split into train and test subsets (90:10) using the “train-test split” from the sklearn library (\cite{Pedregosa2011}). The test set is used only for prediction and validation purposes.
The entire data set is scaled between 0 to 1 using “MinMaxScaler” library of “sklearn” preprocessing module. The primary purpose of scaling is to increase the robustness of the model by taking care of minor standard deviations of features and preserving zero entries in sparse data. The transformation is shown as:
\begin{equation} \label{EQ1} 
	X_{std} =\frac{(X-X_{\min } )}{\left(X_{\max } -X_{\min } \right)}  
\end{equation}
\begin{equation} \label{EQ2} 
	X_{scaled} =X_{std}*(MAX-MIN)+MIN 
\end{equation} 
Where $X_{std}$ and $X_{scaled}$ are the normalized and scaled value corresponding to actual value $X$ .
$X_{max}$ and $X_{min}$ are the maximum and minimum value in the corresponding feature set. $MIN$ and $MAX$ are the feature range to be scaled (in this case $MIN$= 0 and $MAX$= 1).

To avoid over-fitting, “Repeated K-fold” cross validation technique is used. In this current work, number of splits used are ten and repeated over five times. 

\section{Development of MLR and ANN for weld bead geometry parameters}\label{sec5}
\subsection{Linear regression model}\label{subsec5.1}
Linear regression equations are developed for the bead geometry parameters such as the width of the bead, depth of penetration, throat length, and base leg length of the beads created during GMAW process. A mathematical model is developed for predicting the relationship between process parameters and bead geometry. For this current work, three sets of regression equations are established for each output parameter. Preprocessed data obtained from experiments (as shown in Tables \ref{table1} and \ref{table2}) are used for training and testing respectively.
\subsubsection{Multiple linear regression model without interactive terms}\label{subsubsec5.1.1}
First set of regression equation is postulated considering only main effects without considering the interaction effects. A generalized response function for any bead shape parameters can be postulated as:
\begin{equation}\label{eq3}
	{{Y}_{n}}=f(T,I,V,S)
\end{equation}
\begin{equation}\label{eq4}
	{{Y}_{n}}\text{ }=\text{ }{{\alpha }_{0}}+{{\alpha }_{1}}T+{{\alpha }_{2}}I+{{\alpha }_{3}}V+{{\alpha }_{4}}S
\end{equation}

where $Y_{n}$ is the predicted bead geometry parameter and $\alpha _{1},\alpha _{2},\alpha _{3},\alpha _{4}$  are the estimated coefficients.Weld process parameters are denoted in Section \ref{sec2}(Figure \ref{fig:picture1}).

The “linear regression” library of “sklearn” module is applied on the training data set. Coefficients for response
variables are then calculated. Finally, based on linear regression analysis the following linear equations are proposed
as follows:\\

\noindent \textit{Equation for width:}
\begin{align}\label{eq5}
\notag	W&= 5.70231832-0.51518823T+1.28819674I \\
		&-1.20653336V-0.81732366S
\end{align}

\noindent \textit{Equation for depth of penetration:}
\begin{align}\label{eq6}
\notag	P&=1.09927399+0.13730178T+0.95692073I \\
		&+0.78884736V-0.80878598S
\end{align}

\noindent \textit{Equation for length of the throat of the bead:}
\begin{align}\label{eq7}
\notag	TH&=4.57137386-0.75246881T+1.46922465I \\
		&+0.69473209V-1.49694922S
\end{align}

\noindent \textit{Equation for length of the leg base of the bead:}
\begin{align}\label{eq8}
\notag	L&=4.48675941+0.11731523T+1.23693704I \\
		&+0.27055171V-1.0234384S
\end{align}

\subsubsection{Multiple linear regression model using interactive terms}

In this set of regression models, only the interactive terms and the linear terms are considered. When two or more features/variables are combined, they have a substantially larger effect on a feature than when the individual variables are considered alone. This interaction effect is critical to comprehend in regression since we are attempting to investigate the impact of multiple variables on a single response variable.
The extended equation considering interaction effect for any output parameter is formulated as:
\begin{align}\label{eq9}
	{{Y}_{n}}=f(T,I,V,S,TI,TV,TS,IV,IS,VS)	
\end{align}
\begin{align}\label{eq10}
\notag{{Y}_{n}}&={{\alpha}_{0}}+{{\alpha}_{1}}T+{{\alpha}_{2}}I+{{\alpha}_{3}}V+{{\alpha}_{4}}S\text{+}{{\alpha}_{5}}TI \\
\notag			& +{{\alpha }_{6}}TV+{{\alpha }_{7}}TS\text{+}{{\alpha }_{8}}IV \\
				&+{{\alpha }_{9}}IS+{{\alpha }_{10}}VS
\end{align}
where $Y_{n}$ is the predicted bead geometry parameter and ${{\alpha }_{0}},{{\alpha }_{1}}\cdots \text{, }{{\alpha }_{10}}$  are the estimated coefficients.

\noindent \textit{Equation for width:}
\begin{align}\label{eq11}
\notag W&=4.61893649+0.651438T+9.478483I \\ 
\notag	&-1.06198V-3.518106S+0.203403TI \\ 
\notag	&-0.77942TV-0.79978TS-8.114237IV \\
		&-5.8241332IS+9.415024VS
\end{align}

\noindent \textit{Equation for depth of penetration:}
\begin{align}\label{eq12}
	\notag P&=1.04575802+0.308362T+1.717664I \\ 
	\notag	&+1.260269V-2.082993S-0.13729TI\\  
	\notag	&+0.147812TV-0.37407TS-2.024897IV \\ 
			&+1.227235IS+1.06853VS
\end{align}

\noindent \textit{Equation for length of the throat of the bead:}
\begin{align}\label{eq13}
	\notag TH&=5.96133052+0.316904T+1.370493I \\ 
	\notag	&-0.96561V-8.58598S-1.73132TI\\  
	\notag	&-0.00281TV+0.070876TS-2.55739IV \\ 
			&+4.62787IS+6.142991VS
\end{align}

\noindent \textit{Equation for length of the leg base of the bead:}
\begin{align}\label{eq14}
	\notag L&=3.23814441-0.0967T+1.97785I \\
	\notag	&+2.338584V+3.2733419S-0.44175TI \\
	\notag	&+0.593751TV-0.14184TS+0.1332088IV \\
			&-1.080466IS-4.98726VS
\end{align}

\subsubsection{Multiple linear regression model using all terms}
All the higher-order terms, including linear, interactive and quadratic terms are considered in this model. The equation for any output parameters is formulated as :
\begin{align}\label{eq15}
	\notag	{{Y}_{n}}=f(&T,I,V,S,{{T}^{2}},TI,TV,TS,{{I}^{2}}, \\ 
						&IV,IS,{{V}^{2}},VS,{{S}^{2}})
	\end{align}
\begin{align}\label{eq16}
	\notag	{{Y}_{n}}&={{\alpha }_{0}}+{{\alpha }_{1}}T+{{\alpha }_{2}}I+{{\alpha }_{3}}V+{{\alpha}_{4}}S+{{\alpha}_{5}}{{T}^{2}} \\ 
		\notag		& +{{\alpha }_{6}}TI+{{\alpha }_{7}}TV+{{\alpha }_{8}}TS+{{\alpha }_{9}}{{I}^{2}}+{{\alpha }_{10}}IV \\ 
					& +{{\alpha }_{11}}IS+{{\alpha }_{12}}{{V}^{2}}+{{\alpha }_{13}}VS+{{\alpha }_{14}}{{S}^{2}}
\end{align}
Where ${{Y}_{n}}$ is the predicted bead geometry parameter; ${{\alpha }_{0}},{{\alpha }_{1}}\cdots \text{, }{{\alpha }_{14}}$ are the estimated coefficients.

\noindent \textit{Equation for width:}
\begin{align}\label{eq17}
	\notag	W&=4.4932921+0.4135383T+9.08740111I \\
	\notag	&-0.56480994V-1.94290995S+0.49076616{T}^2 \\
	\notag	&-0.42880857TI-0.68836508TV-0.6462735TS \\
	\notag  &+0.09007508{I}^2-8.07186027IV-4.6289759IS \\
	\notag	&-0.52120768{V}^2+9.35456595VS \\
			&-2.21299079{S}^2
\end{align}

\noindent \textit{Equation for depth of penetration:}
\begin{align}\label{eq18}
	\notag	P&=1.04560761+0.57014597T+2.50137859I \\
	\notag	&+1.466591V-2.2697287S-0.76389532{T}^2 \\
	\notag	&+0.79393758TI+0.43326392TV-0.88051647TS \\
	\notag  &-1.19742114{I}^2-2.61559246IV+2.19579487IS \\
	\notag	&-0.28596557{V}^2+1.60877736VS \\
			&-0.53011746{S}^2
\end{align}

\noindent \textit{Equation for length of the throat of the bead:}
\begin{align}\label{eq19}
	\notag	TH&=6.09520908+0.42603142T+2.30330926I \\
	\notag	&-1.32239399V-10.01496405S-0.54508665{T}^2 \\
	\notag	&-0.86131793TI+0.0634657TV-0.20033023TS \\
	\notag  &-1.0122627{I}^2-2.84637547IV+4.417332977IS \\
	\notag	&+0.2734138{V}^2+6.54473054VS \\
			&+1.34339086{S}^2
\end{align}

\noindent \textit{Equation for length of the leg base of the bead:}
\begin{align}\label{eq20}
	\notag	L&=3.2290114+0.13111397T+2.32618553I \\
	\notag	&+2.59597395V+2.98450153S-0.49374791{T}^2 \\
	\notag	&+0.08748134TI+0.72048353TV-0.42511743TS \\
	\notag  &-0.43227202{I}^2-0.15924306IV-0.9190801IS \\
	\notag	&-0.27437132{V}^2-4.74497102VS \\
			&+0.12879011{S}^2
\end{align}
\begin{table*}[h!]
	\centering
	\caption{Statistical table for MLR models }
	\label{table3}
	\resizebox{\textwidth}{!}{%
		\begin{tabular}{ccccccccccccc}
			\hline
			\multirow{2}{*}{Parameters} & \multicolumn{4}{c}{Linear terms} & \multicolumn{4}{c}{Interactive terms} & \multicolumn{4}{c}{All terms} \\ \cline{2-13} 
			& $R^{2}$ & RMSE\footnotemark & STD(test)\footnotemark & STD(model) & $R^{2}$ & RMSE & STD(test) & STD(model) & $R^{2}$ & RMSE & STD(test) & STD(model) \\ \hline
			Width & 0.825 & 0.182 & 0.484 & 0.379 & 0.9905 & 0.0673 & 0.484 & 0.4703 & 0.9884 & 0.0770 & 0.484 & 0.5014 \\
			Penetration & 0.969 & 0.062 & 0.248 & 0.255 & 0.9199 & 0.1006 & 0.248 & 0.2029 & 0.9697 & 0.0673 & 0.248 & 0.2114 \\
			Throat & 0.432 & 0.251 & 0.254 & 0.210 & 0.9585 & 0.0746 & 0.254 & 0.2274 & 0.9483 & 0.0822 & 0.254 & 0.2268 \\
			Leg & 0.724 & 0.182 & 0.249 & 0.238 & 0.9793 & 0.0542 & 0.249 & 0.2640 & 0.9823 & 0.0524 & 0.249 & 0.2689 \\ \hline
		\end{tabular}%
	}
\end{table*}
\begin{table*}[h!]
	\centering
	\caption{Statistical table for ANN models }
	\label{table4}
	\resizebox{\textwidth}{!}
	{%
		\begin{tabular}{ccccccccccccc}
			\hline
			\multirow{2}{*}{Parameters} & \multicolumn{4}{c}{Linear terms} & \multicolumn{4}{c}{Interactive terms} & \multicolumn{4}{c}{All terms} \\ \cline{2-13} 
			& $R^{2}$ & RMSE & STD(test) & STD(model) & $R^{2}$ & RMSE & STD(test) & STD(model) & $R^{2}$ & RMSE & STD(test) & STD(model) \\ \hline
			Width & 0.979 & 0.099 & 0.484 & 0.451 & 0.9562 & 0.1427 & 0.484 & 0.4466 & 0.9510 & 0.1715 & 0.484 & 0.544 \\
			Penetration & 0.923 & 0.100 & 0.248 & 0.198 & 0.8295 & 0.1388 & 0.248 & 0.2132 & 0.9079 & 0.1043 & 0.248 & 0.2177 \\
			Throat & 0.908 & 0.106 & 0.254 & 0.229 & 0.8849 & 0.1191 & 0.254 & 0.2357 & 0.9001 & 0.1208 & 0.254 & 0.2771 \\
			Leg & 0.942 & 0.087 & 0.249 & 0.257 & 0.9222 & 0.0989 & 0.249 & 0.2092 & 0.9535 & 0.1059 & 0.249 & 0.3126 \\ \hline
		\end{tabular}%
	}
\end{table*}
Table \ref{table3} shows the detailed statistics of the performance of this model. It can be noticed that the R-squared values of the regression relations improved as the degree of correlation increased. For first sets of regression relations, the R-squared value (0.432) for throat length of the bead geometry parameter was poor. This value is refined significantly in the second and third sets of regression relations (0.9585 and 0.9483) where higher order and interaction terms are considered.

\subsection{Modeling of the Artificial Neural Network}
An artificial neural network is a tool to predict complicated relations between input and output parameters (\cite{Malinov2004}), mimicking just a biological neural network. These neural networks can learn from the process and predict the solution of complex problems with great accuracy. Due to low prediction error, ANN is preferable over regression analysis for the same data set. A detailed comparative analysis of bead geometry in submerged arc welding is provided by \cite{Sarkar2016}.

\begin{figure}[!h]
	\centering
	\includegraphics[width=0.5\linewidth]{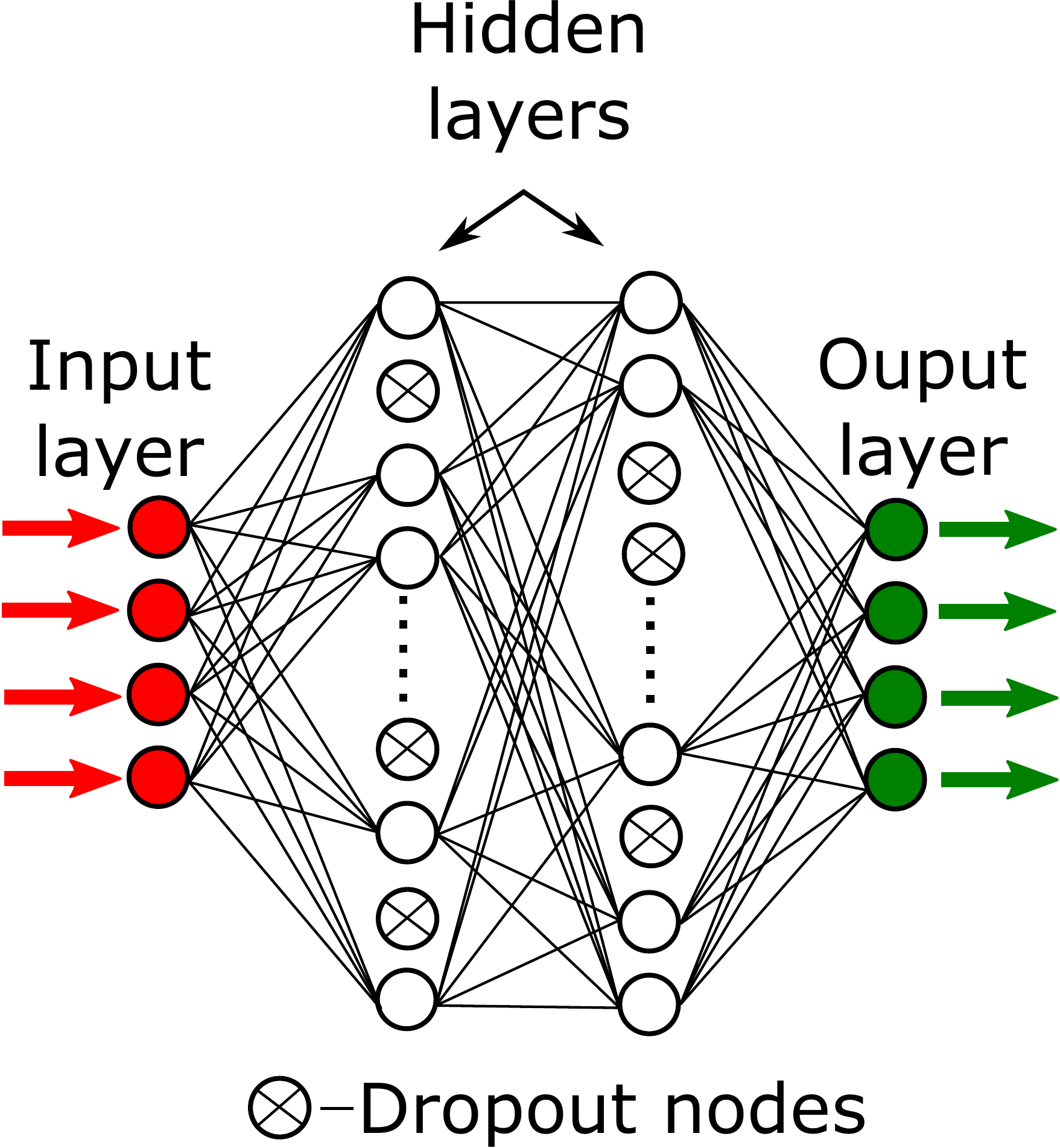}
	\caption{Artificial neural network schematic}
	\label{fig:picture5}
\end{figure}

The “Tensorflow” module is implemented to design the neural network for this study. The primary architecture of the neural network model is emulated in Figure \ref{fig:picture5}. The neural network consists of different layers and neurons. Neurons are the nodes through which the information propagates forward and backward from one layer to the next layer. This model consists of an input layer (which receives input parameters through a nodes), the hidden layers, and an output layer (for each of the predictions of response variables). After preprocessing the data set as explained in section \ref{sec4}, it is further processed for modeling and training.
\subsubsection{Hyper-parameter tuning}
Hyper-parameters are the ANN model parameters that dictate the model performance. The hyper-parameters tuned for this study are the number of hidden layers, dropout rate, and learning rate. “KerasTuner” library package is used for the tuning process (\cite{omalley2019kerastuner}). The “Random search” method is used for tuning the hyper-parameters. Multiple dense hidden layers are taken (with iterating neurons between 30 to 40 with step size =1). The range for dropout rate is selected from 0.1 to 0.6 in a step of 0.1. The learning rate for the model is kept between 0.01 and 0.02. The objective is to minimize the loss function (mean absolute error). Hence, the maximum trials are taken as 100 with 2 executions per trial, and it is run for 2500 epochs.
\begin{figure}[h!]
	\centering
	\includegraphics[width=0.7\linewidth]{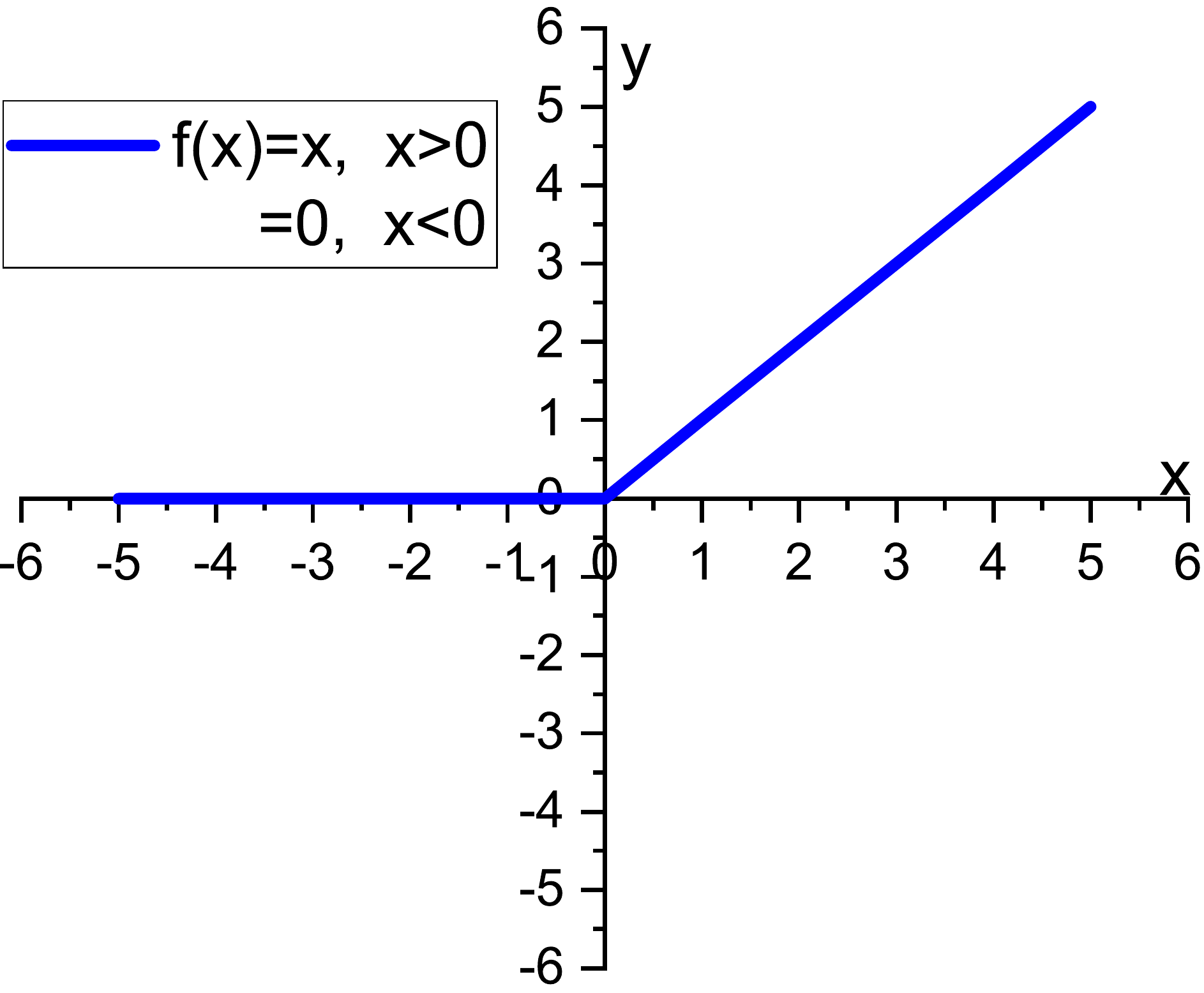}
	\caption{ReLu activation function}
	\label{fig:picture6}
\end{figure}
\subsubsection{ANN model}
\begin{figure}[h!]
	\centering
	\subfloat[]{%
		\includegraphics[width=0.2\textwidth,height=0.17\textwidth]{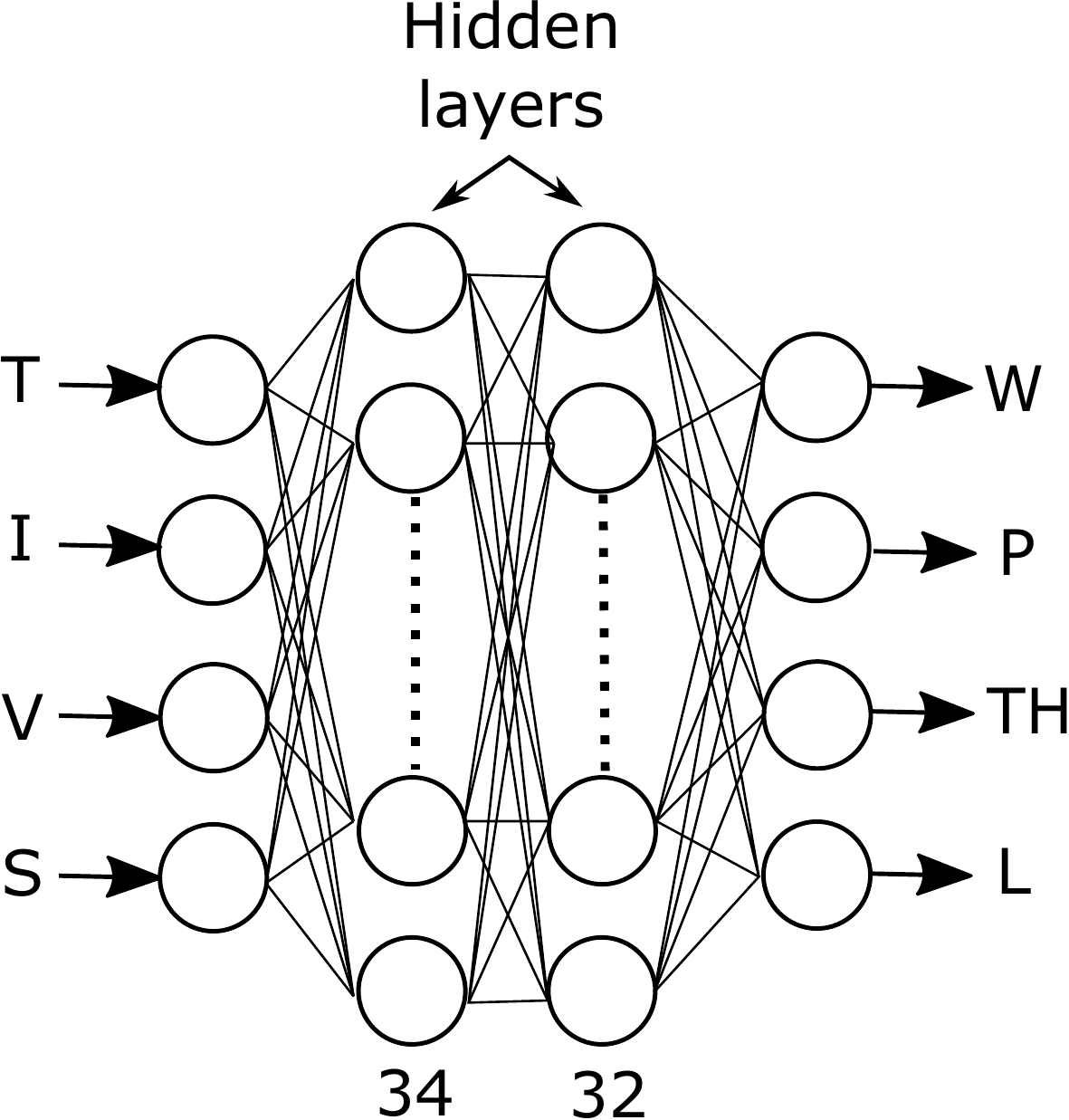}
		\label{fig:picture7a}}
	
	\subfloat[]{
		\includegraphics[width=0.2\textwidth,height=0.17\textwidth]{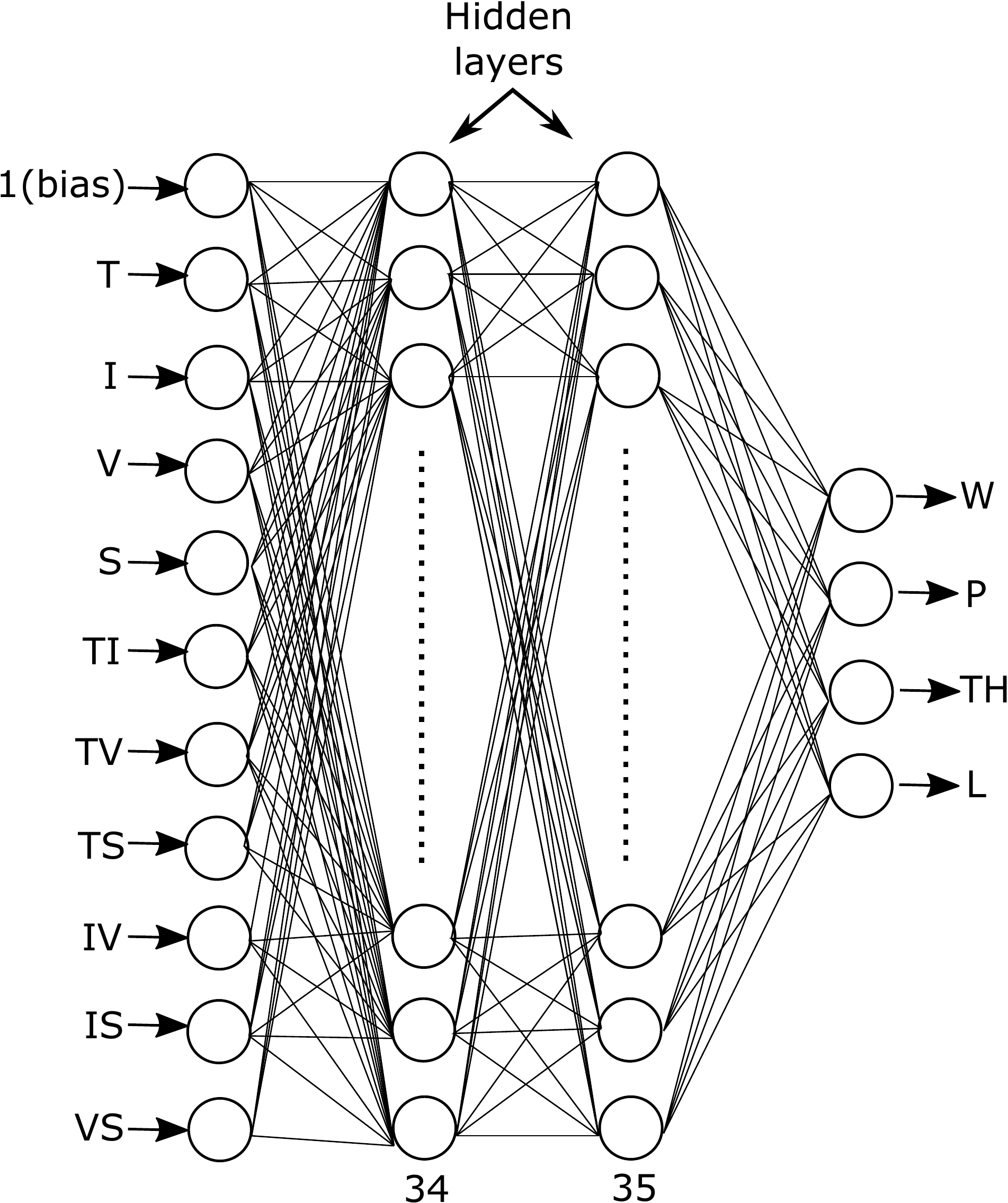}
		\label{fig:picture7b}}
	
	\subfloat[]{
		\includegraphics[width=0.2\textwidth,height=0.17\textwidth]{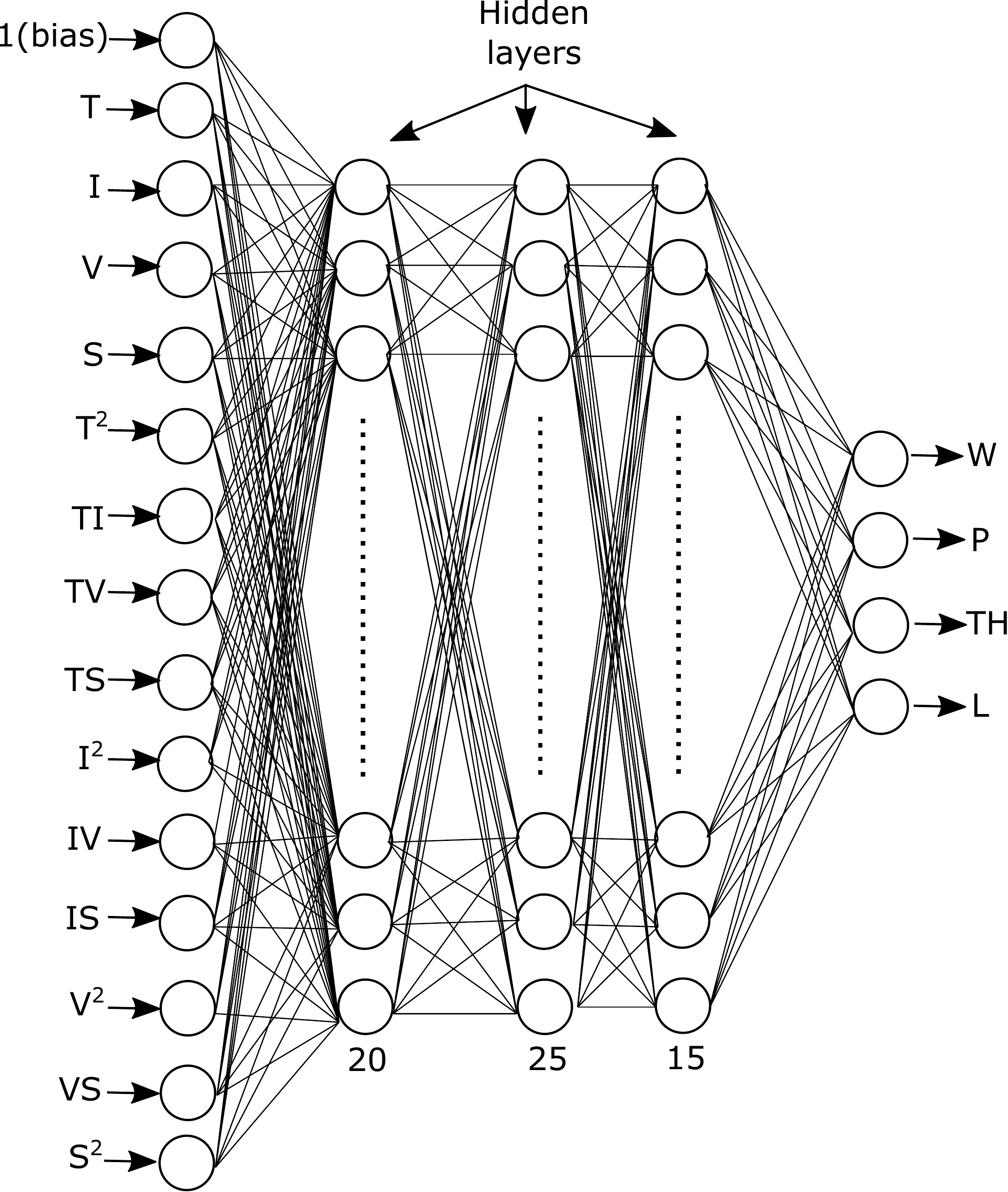}
		\label{fig:picture7c}}
	\caption{ANN models (a) linear terms only (b) linear and interaction terms (c) all terms}
	\label{fig:picture7}
\end{figure}
\footnotetext[1]{RMSE= root mean square error}
\footnotetext[2]{STD= standard deviation}
After fine-tuning hyperparameters, the final ANN model is implemented. Three sets of ANN models are formed. Each model has three types of layers (input, hidden, and output layer), with a dropout layer placed between the hidden layers at a rate of 0.1. The “Rectified Linear Unit activation function (ReLu)” was chosen as the activation function (\cite{Agarap2018}) for hidden layers and “Linear” activation was used for the output layer(Figure \ref{fig:picture6}). Layer weights are initialized randomly using a “he-uniform” initializer (\cite{he2015delving}). A computationally efficient “Adam” optimizer as explained by \cite{kingma2017adam} is used to update weights and monitor the optimization process. 
\begin{figure*}[h!]
	\centering
	\subfloat[]{%
		\includegraphics[width=0.3\linewidth, height=0.25\textwidth]{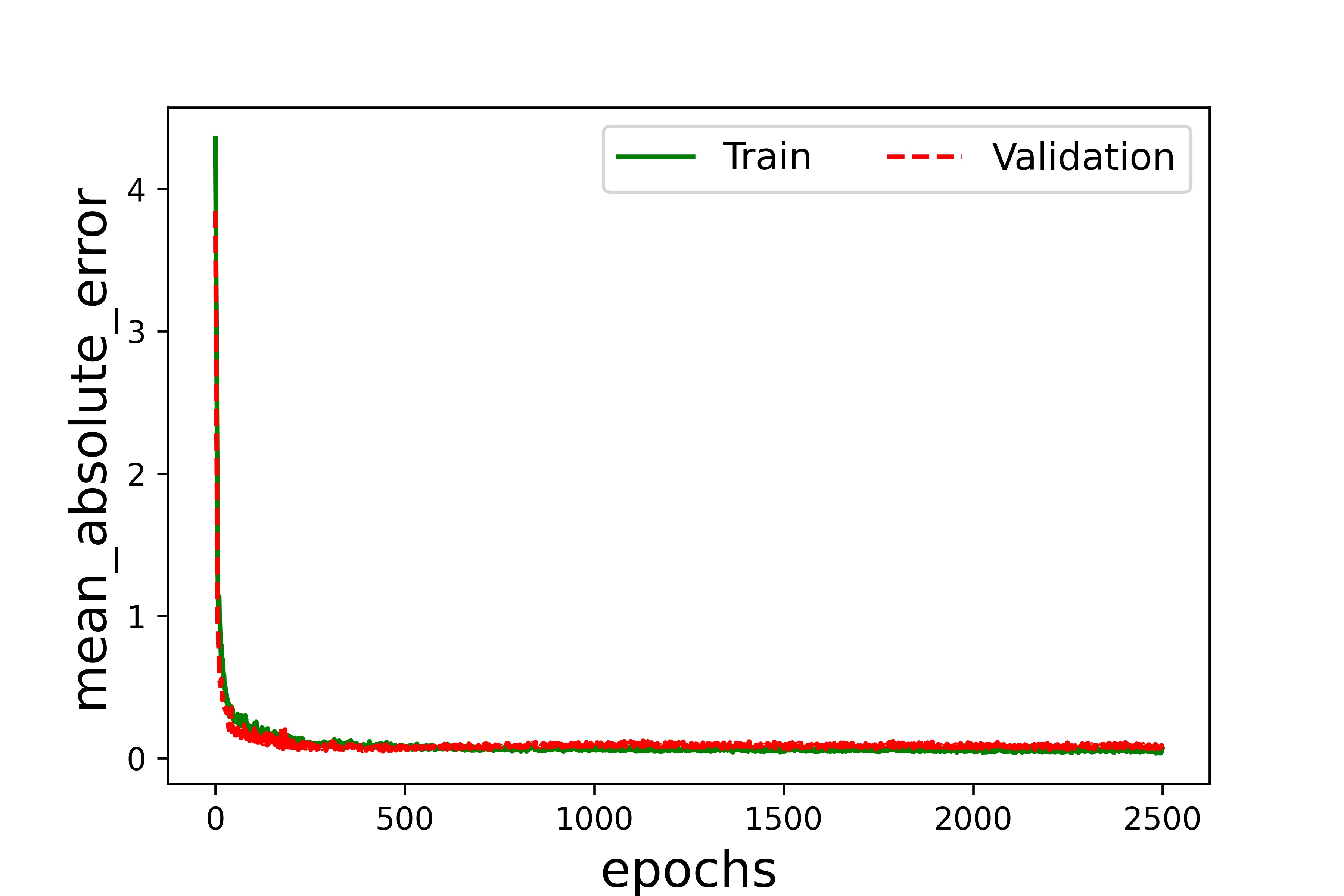}
		\label{fig:picture8a}}
	\subfloat[]{
		\includegraphics[width=0.3\linewidth, height=0.25\textwidth]{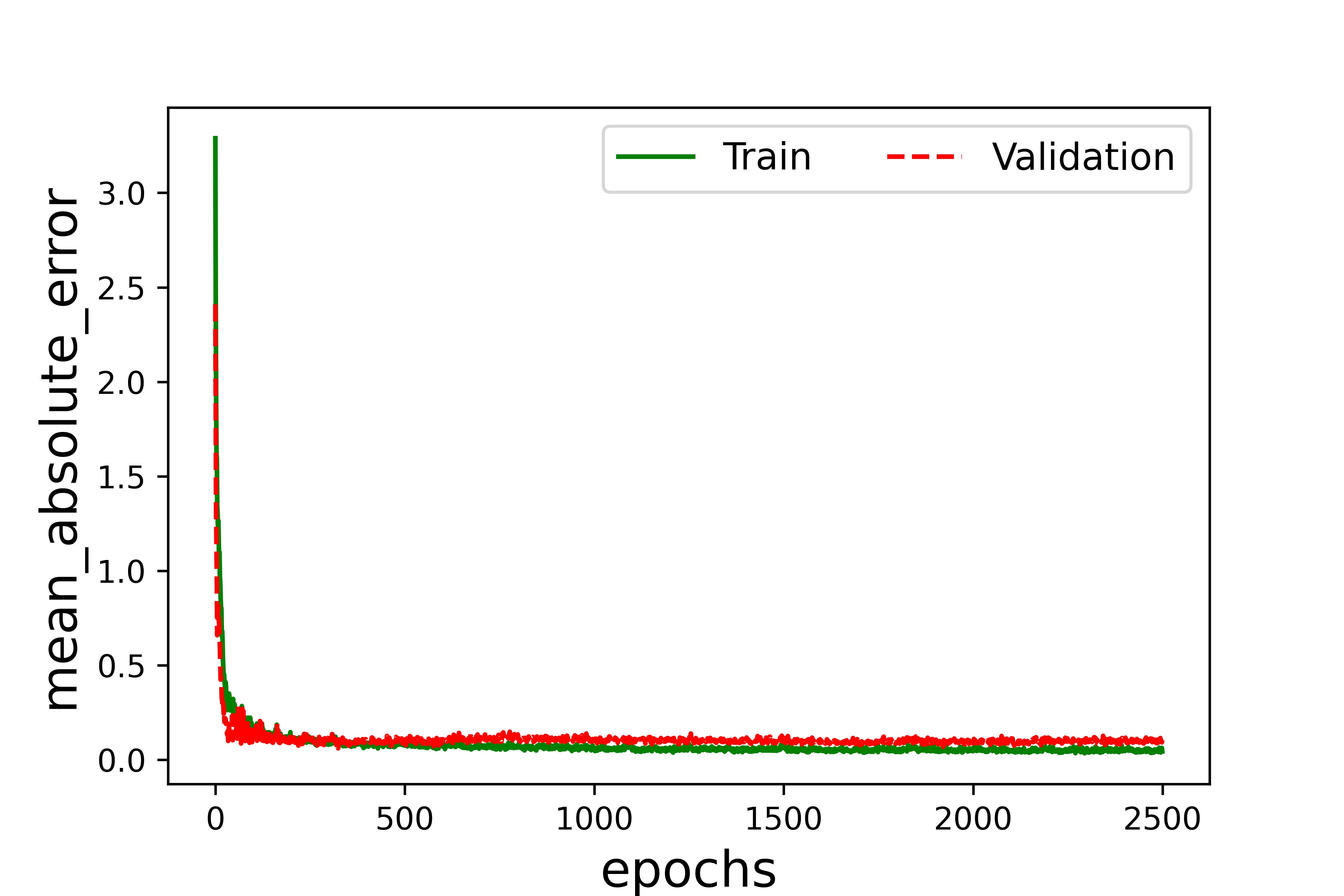}
		\label{fig:picture8b}}
	\subfloat[]{
		\includegraphics[width=0.3\linewidth, height=0.25\textwidth]{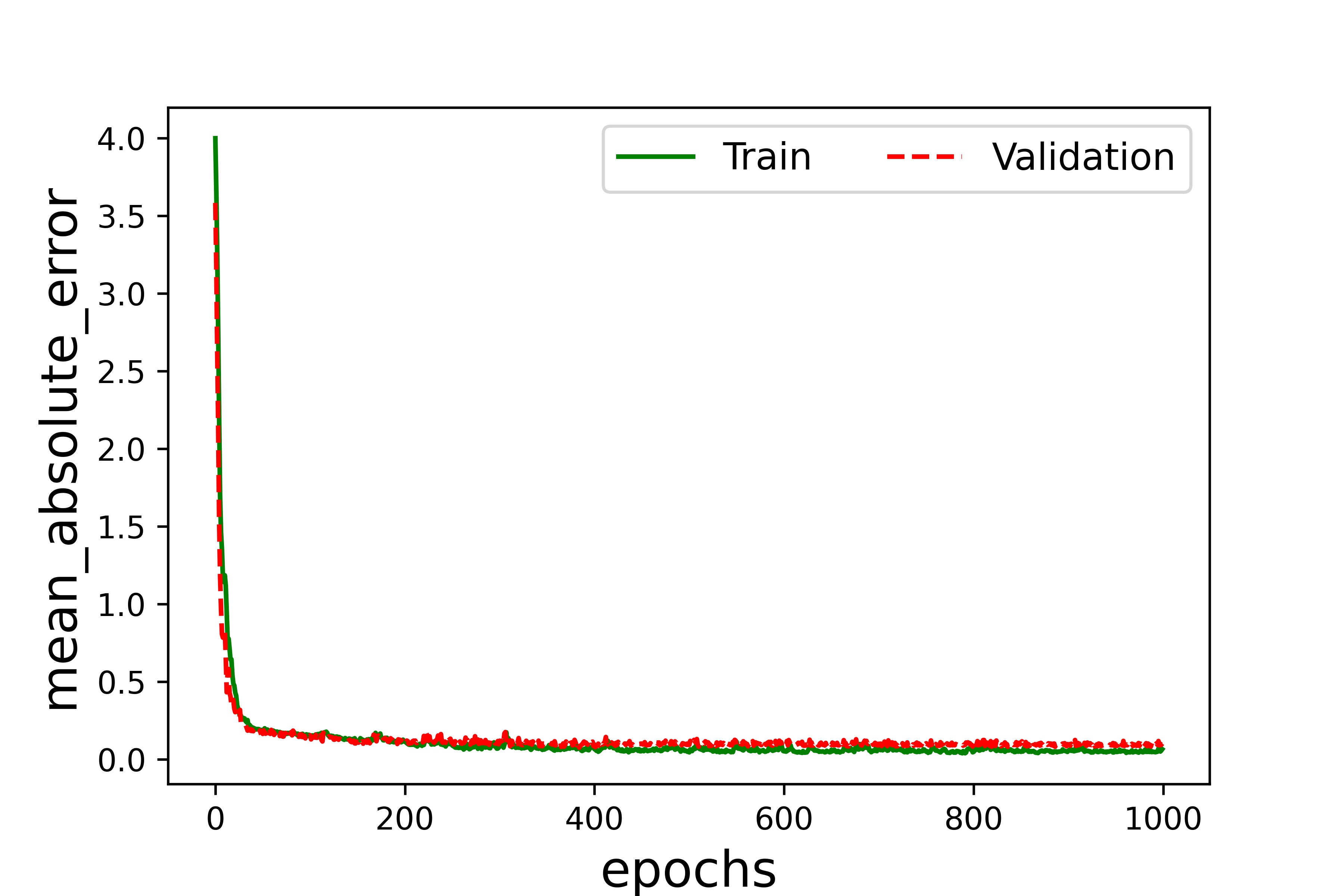}
		\label{fig:picture8c}}
	\caption{Neural network performance plot for ANN models (a)with linear terms (b) with interactive terms (c) with all terms}
	\label{fig:picture8}
\end{figure*}
Figure \ref{fig:picture7a} shows the first set of ANN model considering linear terms only. The first hidden layer has 34 neurons and the second hidden layer has 32 neurons. Figure \ref{fig:picture7b} explains the second set of ANN model with interactive terms having 11 neurons in input layer.The two hidden layers in this model consists of 34 and 35 neurons respectively. Similarly Figure \ref{fig:picture7c} represents the third set of ANN model including all terms having 15 neurons in input layer. This model consists of three hidden layers with 20,25 and 15 neurons respectively. In all the models a dropout layer is placed in-between hidden layers at a rate of 0.1.
   
Table \ref{table4} shows the statistical data of all the models. It is found that the R-squared values of the depth of penetration and throat parameters decreased slightly in case of ANN model considering interactive terms. Other than this there is no significant improvement in the ANN models as the higher order terms are considered.

Figure \ref{fig:picture8} displays the mean absolute errors over the total number of epochs for the train and the validation data. The mean absolute error between the train and validation data is stabilized after 2000 epochs for first and second case and 800 epochs for third case. The loss function for the train and validation data in Figure \ref{fig:picture8} does not diverge, suggesting the ANN to be a perfect fit for the training process.

%
\subsubsection{Interpretation of ANN model}
To explain the impact of the input variable on the ANN model a unique SHAP (SHapley Additive exPlanations)(as presented by \cite{lundberg2017unified}) model is implemented in the Python script. This algorithm assigns an importance value (SHAP value) for a prediction. Each feature or an input variable has a contribution to the prediction of an output. The SHAP values are broadly understood to be the average of these marginal contributions of individual features over the predictions made by the ANN model. The SHAP value can be expressed as a value function “$S$” of features in “$C$” subset as in the following equation(\cite{lundberg2017unified}). 
\begin{small}
\begin{align}\label{eq21}
	{{\phi }_{k}}(S)=\sum\limits_{\frac{C\subseteq \left\{1,...,n \right\}}{\left\{ k \right\}}}\mathcal{F}\left\{S(C\cup(\left\{k\right\}))-S(C)\right\}
	\end{align}
\end{small}
\begin{small}
\begin{align} \label{eq22} 
	{{S}_{x}}(C)=\int{f({{x}_{1}},....,{{x}_{n}}) d}{{\mathbb{P}}_{x\notin C}}-{{E}_{\mathbf{X}}}(f(\mathbf{X}))
\end{align} 
\end{small}
Where, $\mathcal{F}$=${{\lvert C \rvert!\left( n-\lvert C \rvert-1 \right)!}\slash{n!}}$, ${\phi }_{k}$ is the contribution of $k^{th}$ feature on the prediction model $f\left(\mathbf{X} \right)$ , $C$ is a subset of the feature or input variable, $n$ is the number of features, $\mathbf{X}$  is the vector of feature values of instance. ${S}_{x}\left( C \right)$  is the prediction of the marginalized feature values in set $C$ that are excluded from set $C$ .

After calculating SHAP values for each instance, a matrix of SHAP values is obtained. This matrix has one row per data instance and one column per feature. We can interpret the entire model by analyzing the SHAP values in this matrix.

\section{Results and discussion}
\begin{figure*}[!h]
	\centering
	\subfloat[]{%
		\includegraphics[width=0.3\linewidth, height=0.2\textwidth]{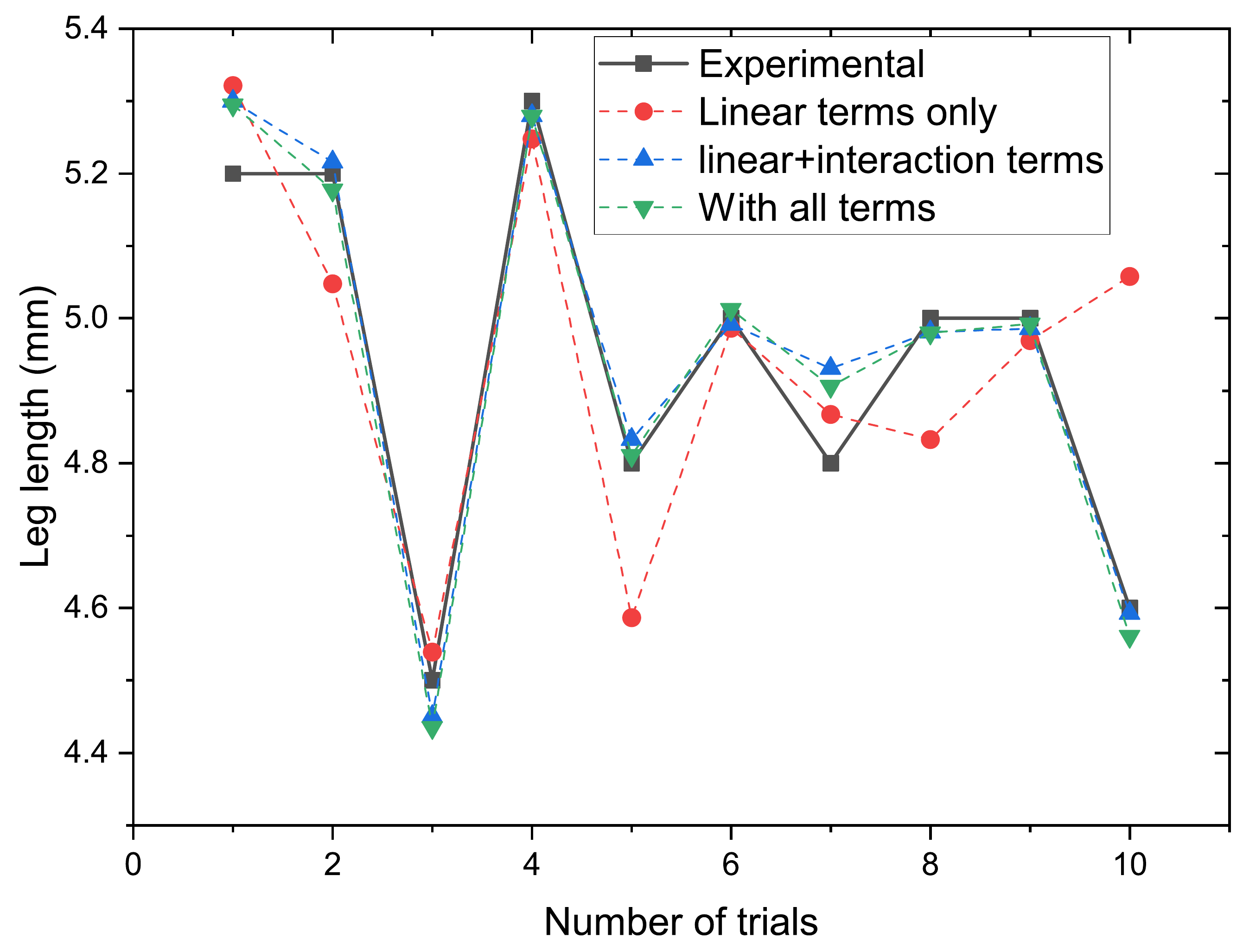}
		\label{fig:picture9a}}
	\subfloat[]{
		\includegraphics[width=0.3\linewidth, height=0.2\textwidth]{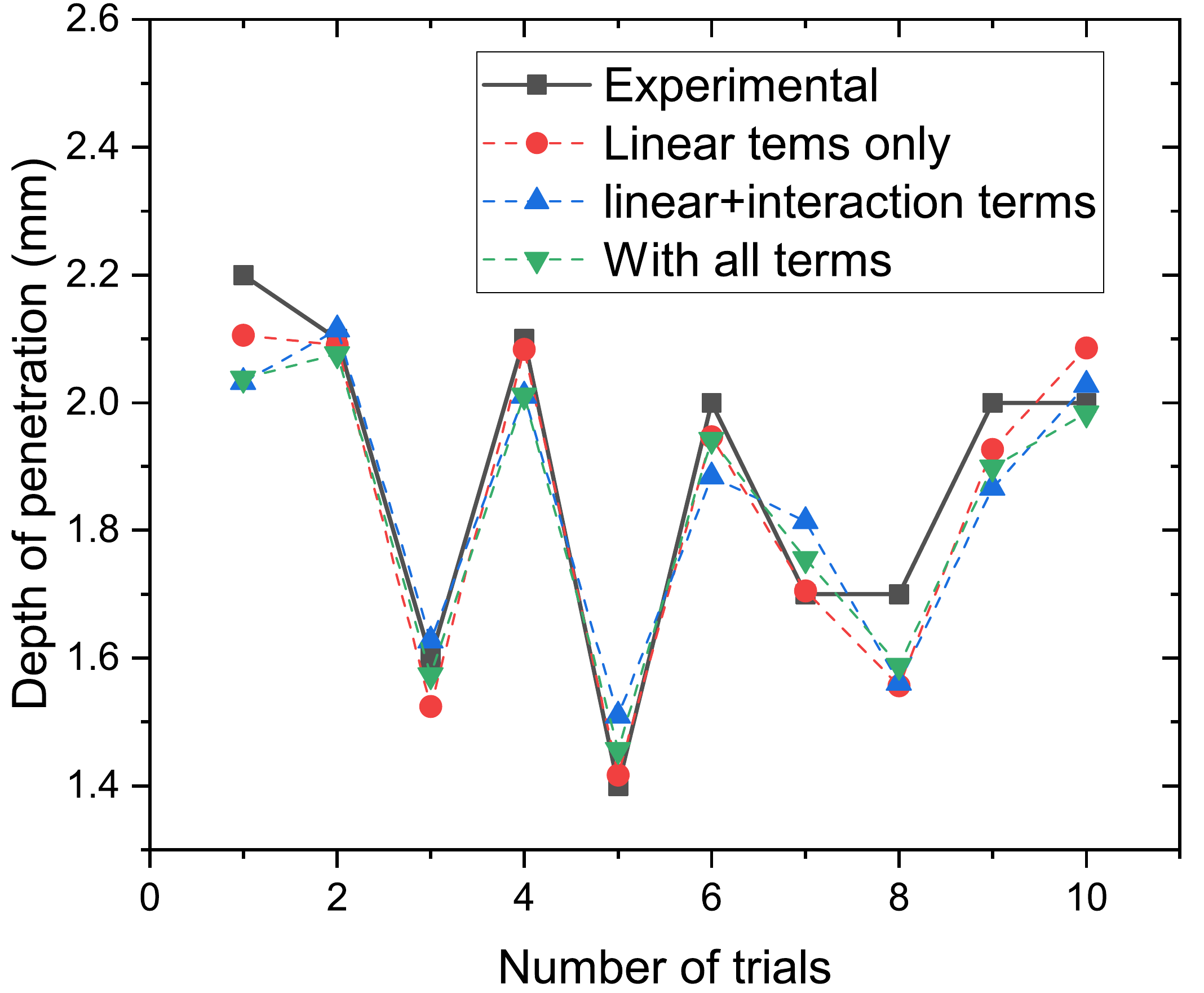}
		\label{fig:picture9b}}
	
	\subfloat[]{
		\includegraphics[width=0.3\linewidth, height=0.2\textwidth]{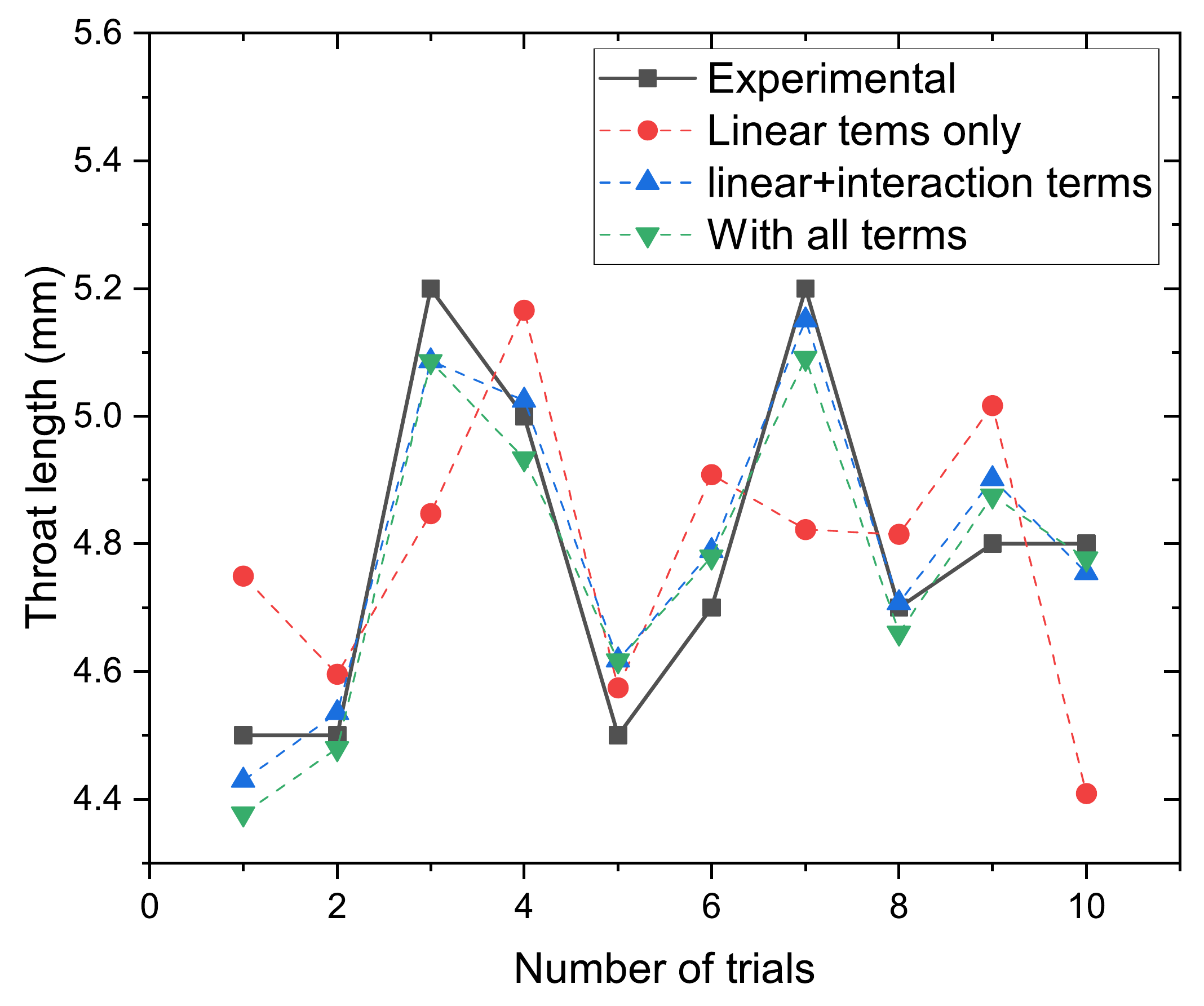}
		\label{fig:picture9c}}
	\subfloat[]{
		\includegraphics[width=0.3\linewidth, height=0.2\textwidth]{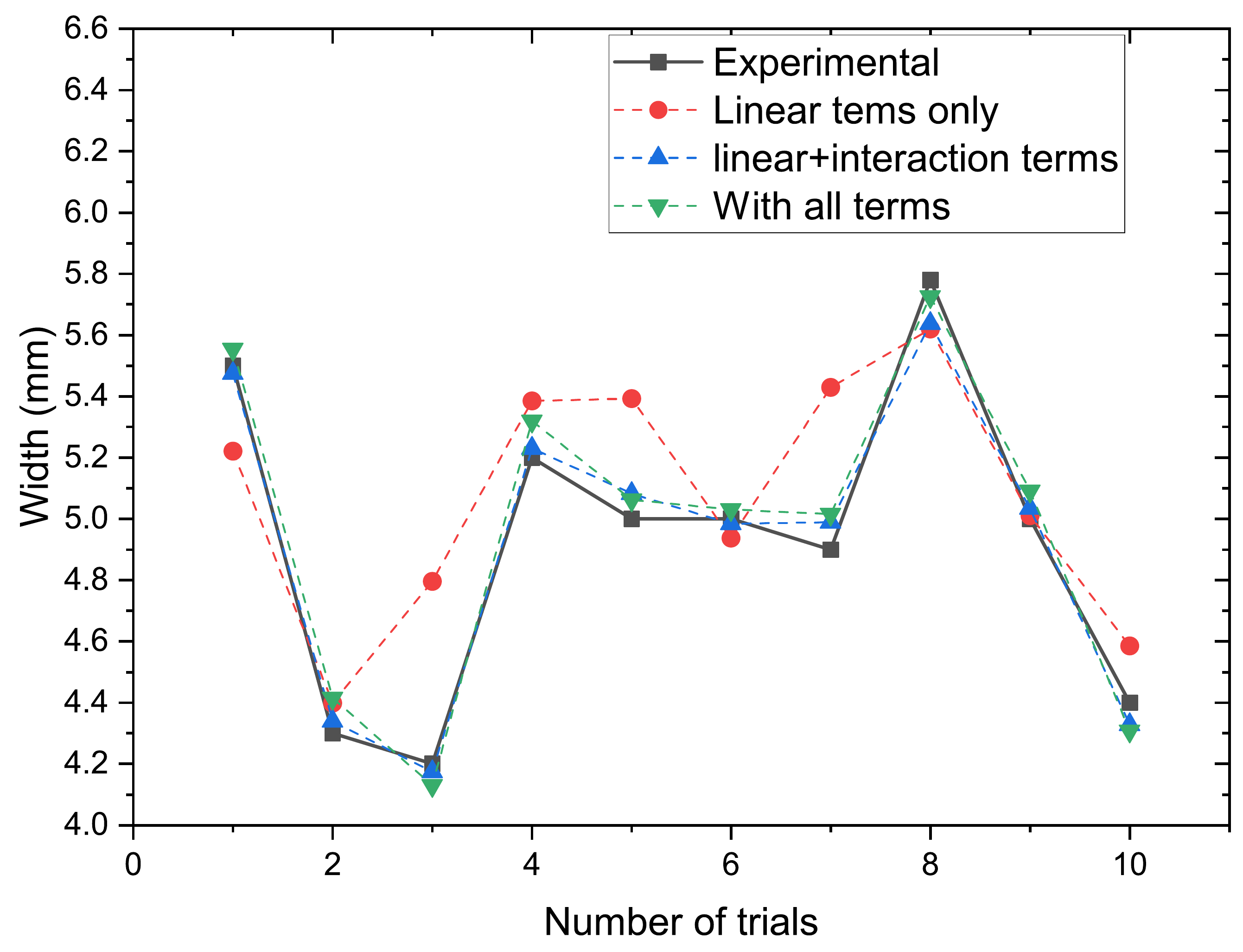}
		\label{fig:picture9d}}
	\caption{Experimental vs predicted values for MLR model}
	\label{fig:picture9}
\end{figure*}
\begin{figure*}[!h]
	\centering
	\subfloat[]{%
		\includegraphics[width=0.3\linewidth, height=0.2\textwidth]{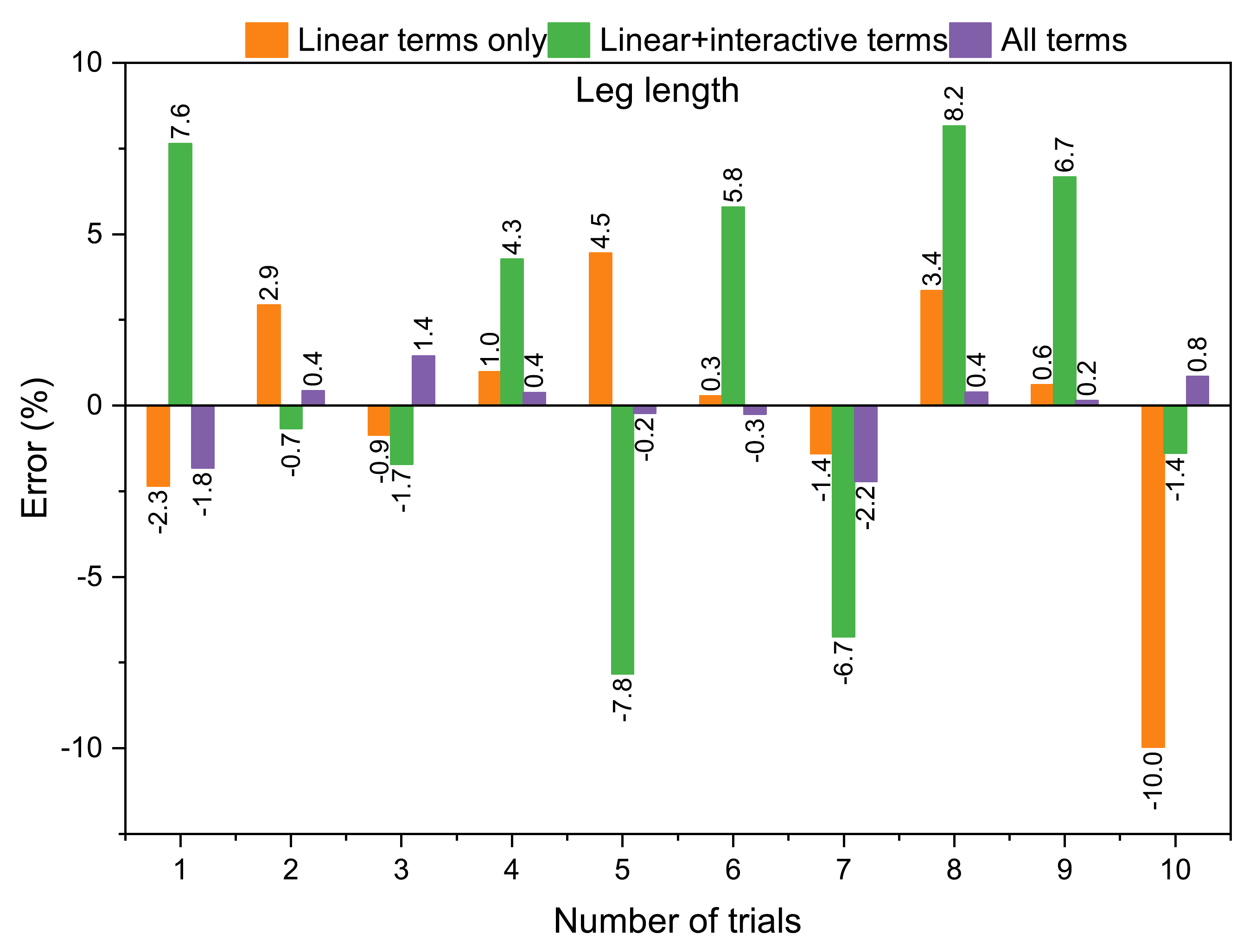}
		\label{fig:picture10a}}
	\subfloat[]{
		\includegraphics[width=0.3\linewidth, height=0.2\textwidth]{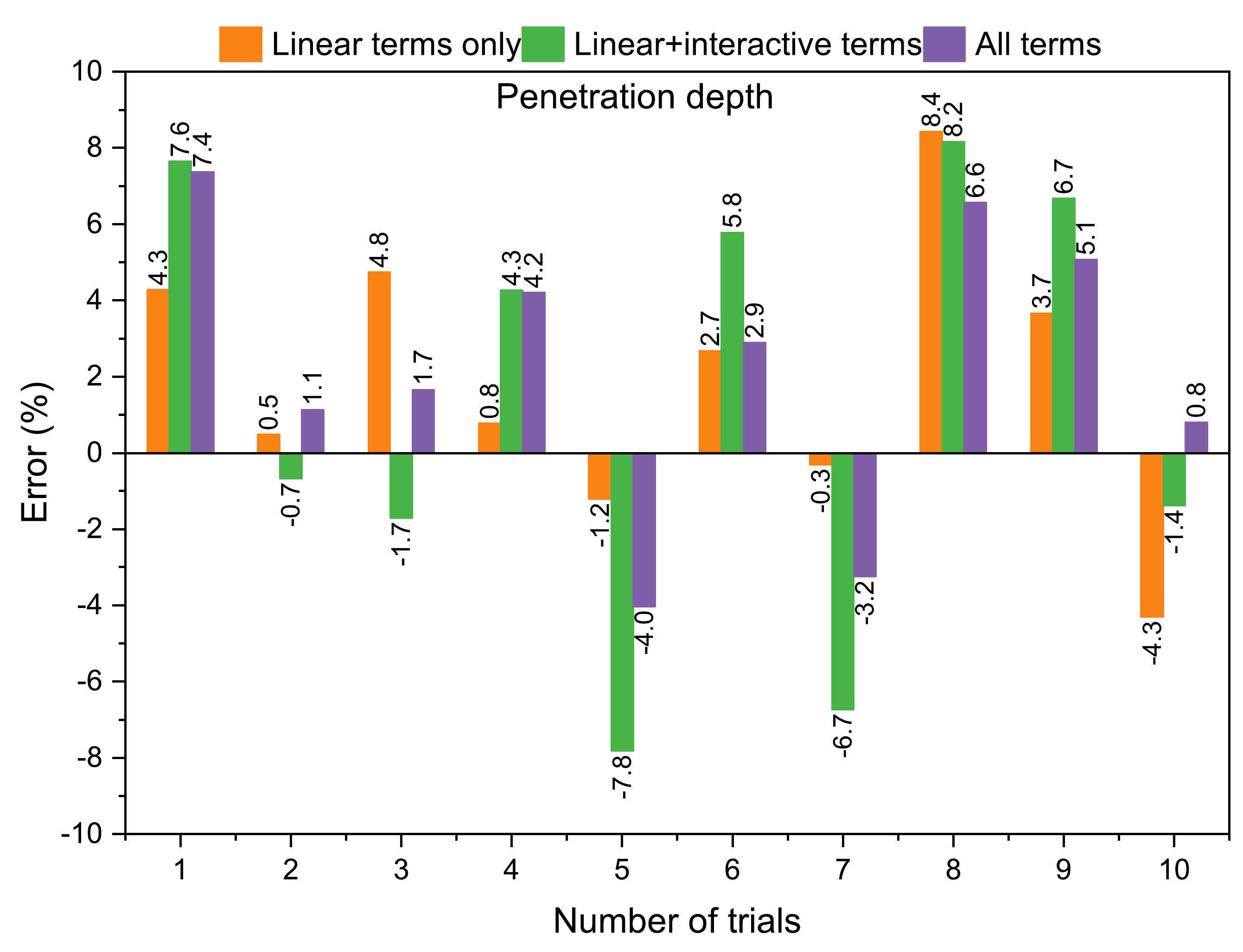}
		\label{fig:picture10b}}
	
	\subfloat[]{
		\includegraphics[width=0.3\linewidth, height=0.2\textwidth]{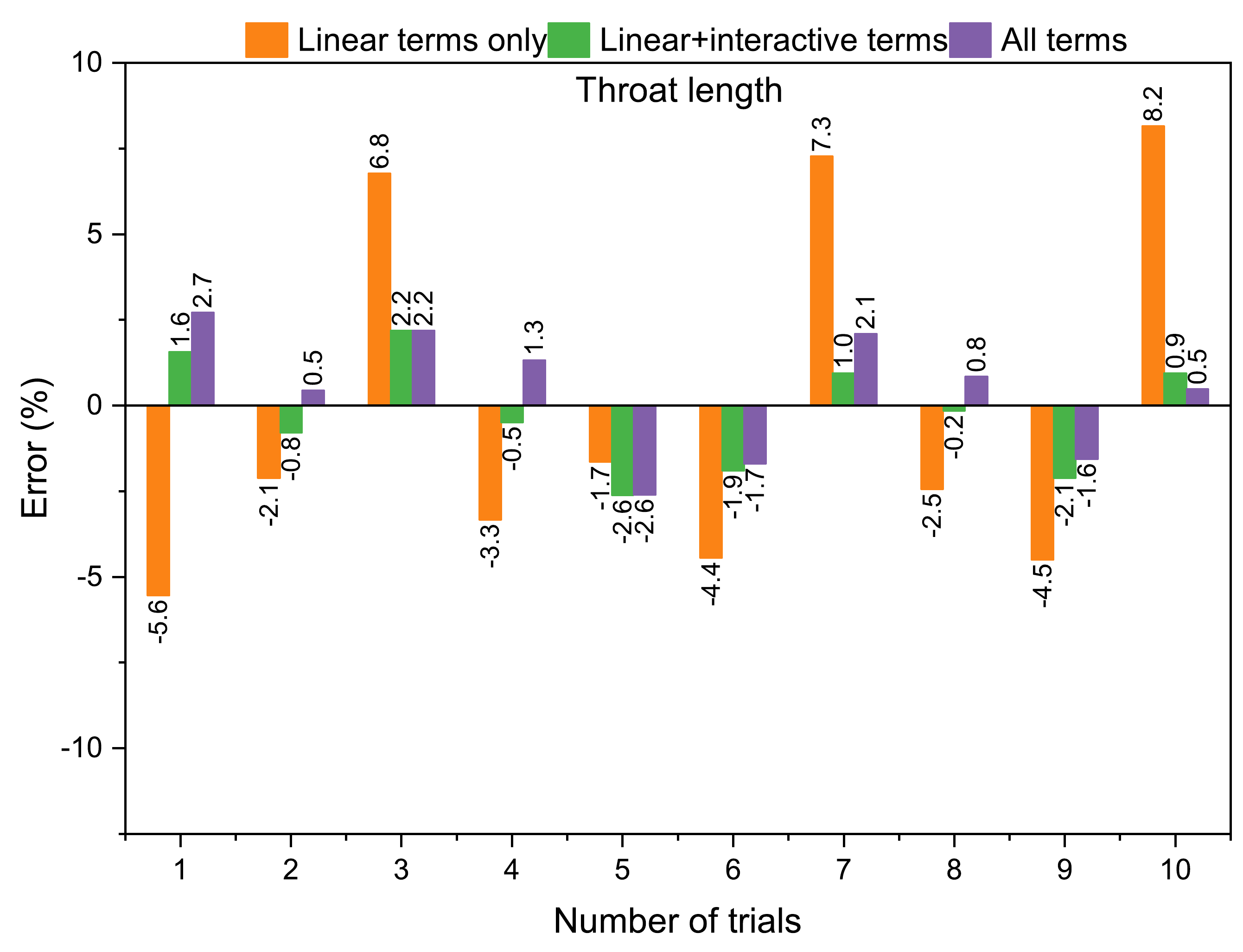}
		\label{fig:picture10c}}
	\subfloat[]{
		\includegraphics[width=0.3\linewidth, height=0.2\textwidth]{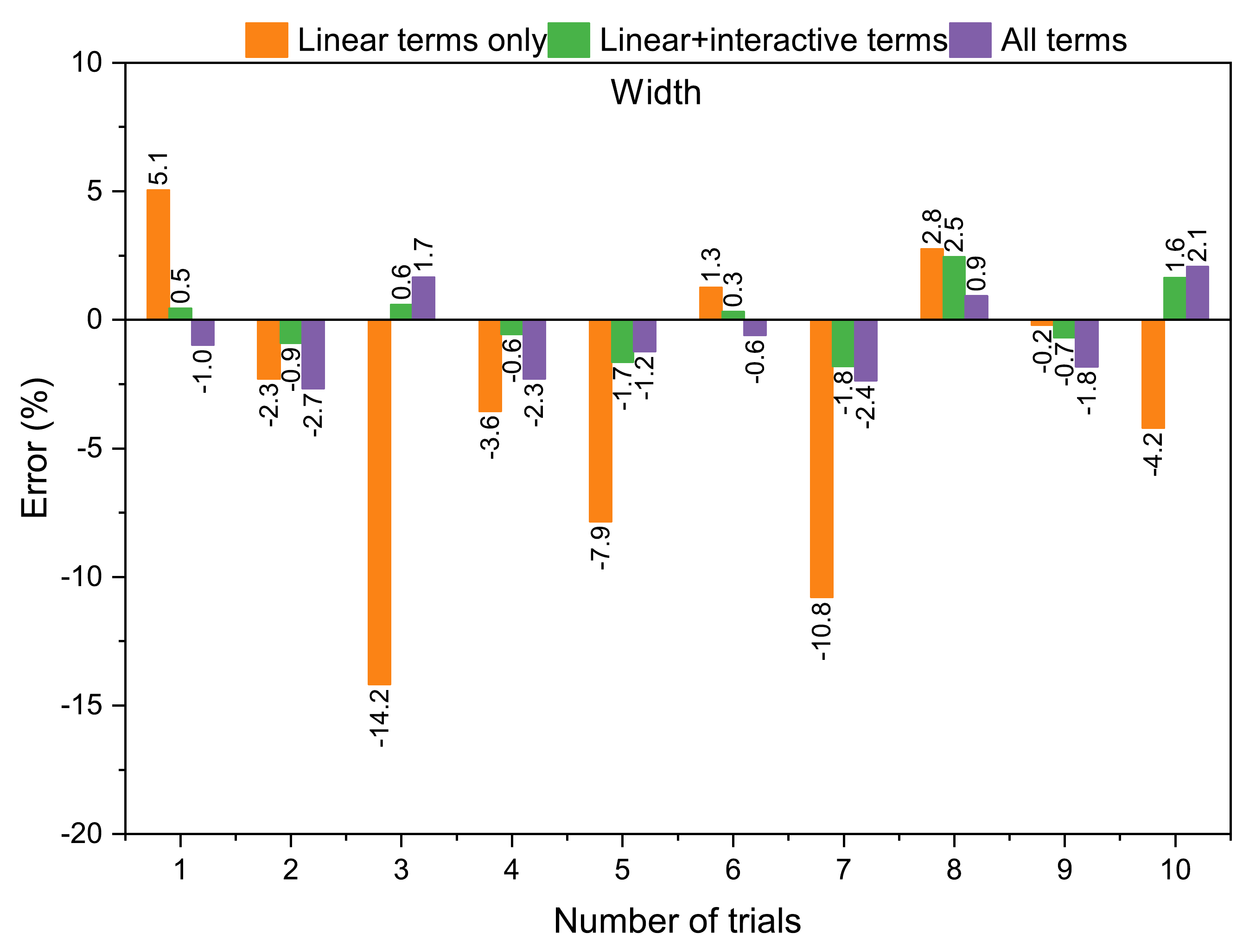}
		\label{fig:picture10d}}
	\caption{Percentage of error in MLR prediction}
	\label{fig:picture10}
\end{figure*}
 As explained earlier,three types of regression models and ANN models are developed to predict the fillet weld bead size. First set of regression equations are developed only by considering linear terms as explained in the equations (\ref{EQ1}-\ref{eq8}). The second set of equations (\ref{eq9}-\ref{eq14}) are modelled using both linear and interactive terms excluding the quadratic terms. In the third set of equations (\ref{eq15}-\ref{eq20}) all terms are considered.

The validity of all the regression models can be verified from the comparison plot provided in the figure \ref{fig:picture9}. Figures \ref{fig:picture9a}-\ref{fig:picture9d} show the comparison between experimental (trial datasets otherwise called as test datasets provided in Table \ref{table2}) and predicted value (obtained from all the three regression models). The plots indicate satisfactory results between the actual and the estimated values of weld bead geometry except for a few combinations.

\begin{figure*}[tph]
	\centering
	\subfloat[]{%
		\includegraphics[width=.4\linewidth, height=0.3\textwidth]{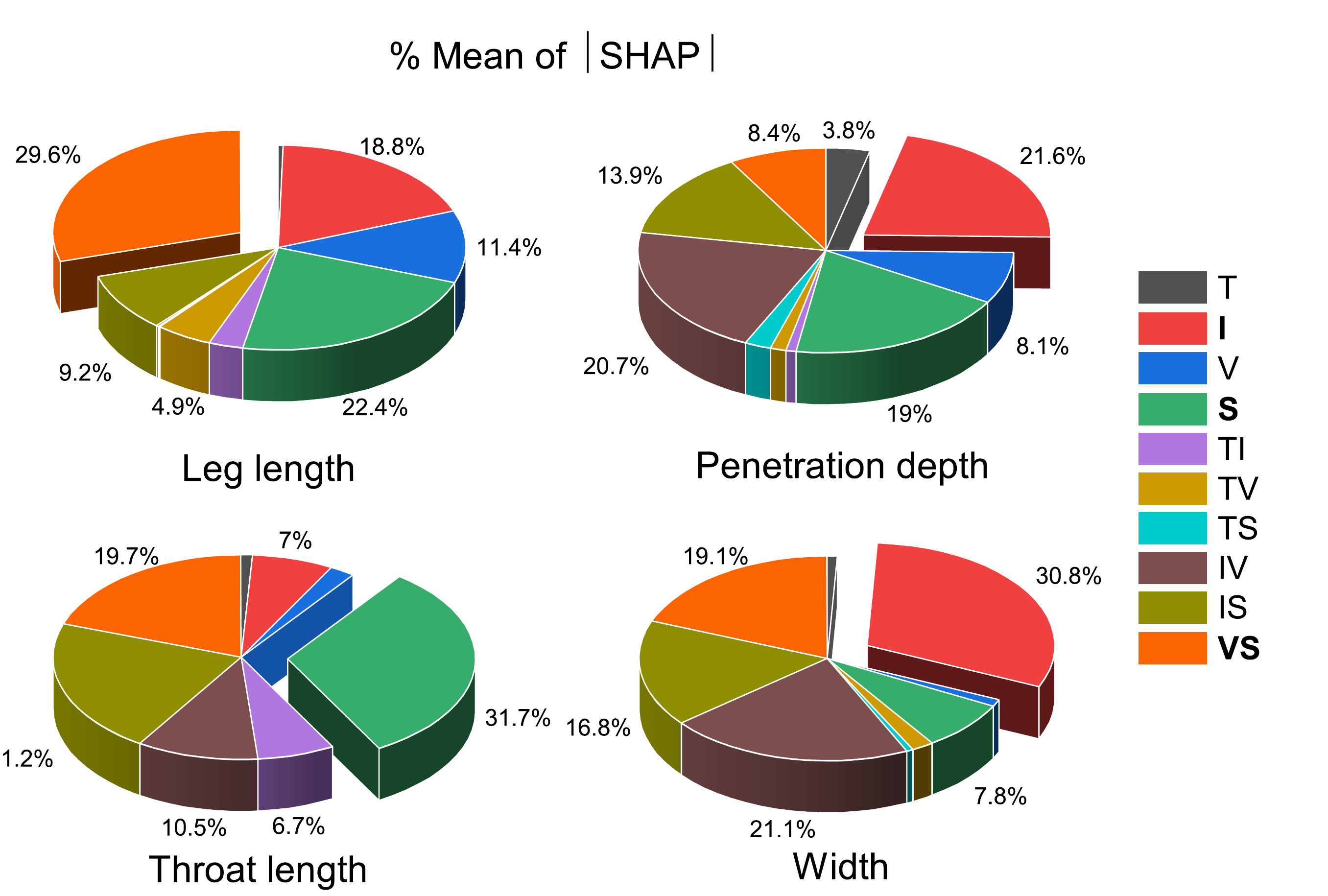}
		\label{fig:picture11a}}
	\subfloat[]{
		\includegraphics[width=0.4\linewidth, height=0.3\textwidth]{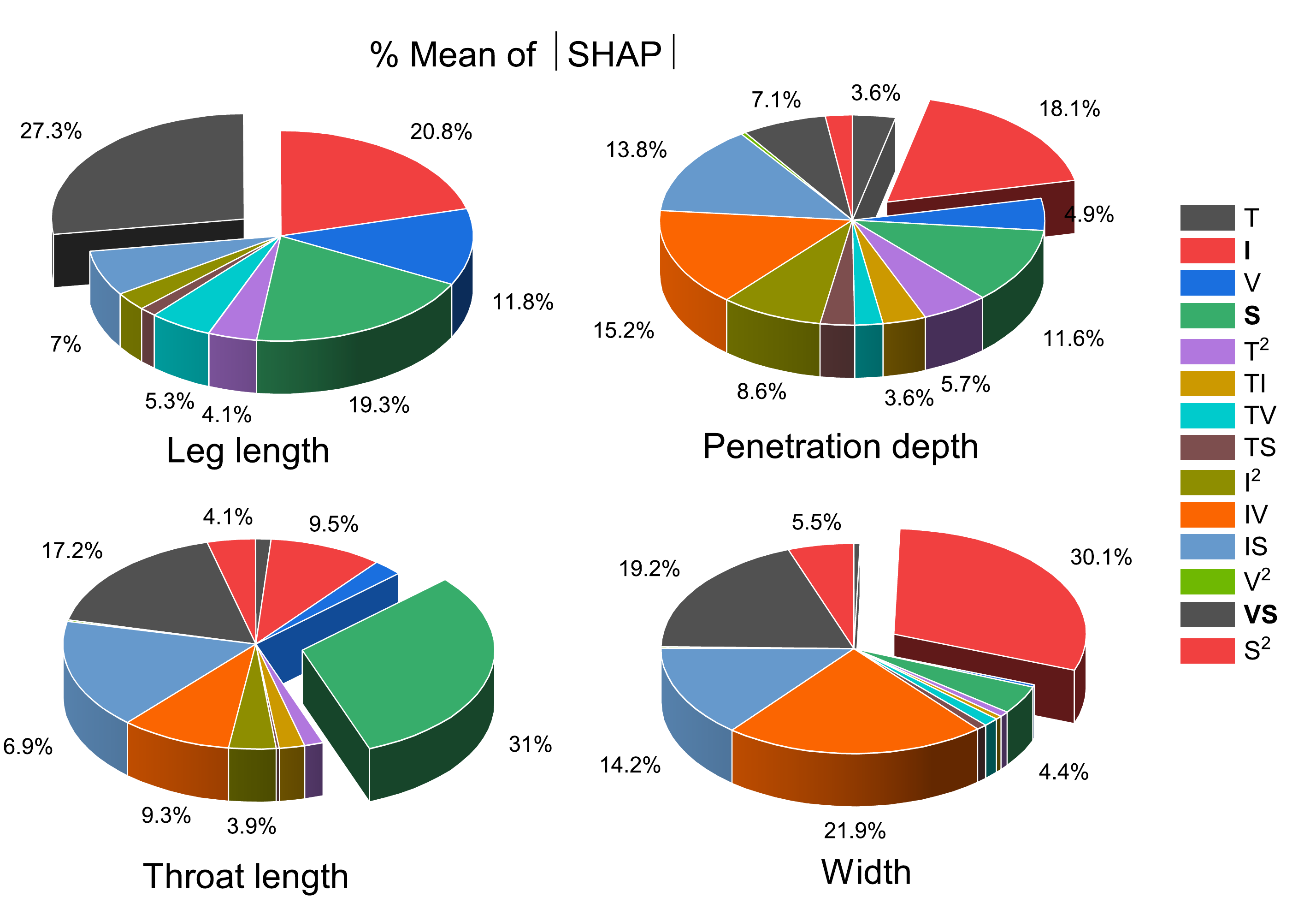}
		\label{fig:picture11b}}
	\caption{Feature importance on MLR model(a) with linear and interactive terms (b) with all terms}
	\label{fig:picture11}
\end{figure*}
\begin{figure*}[tph]
	\centering
	\subfloat[]{%
		\includegraphics[width=0.3\linewidth, height=0.2\textwidth]{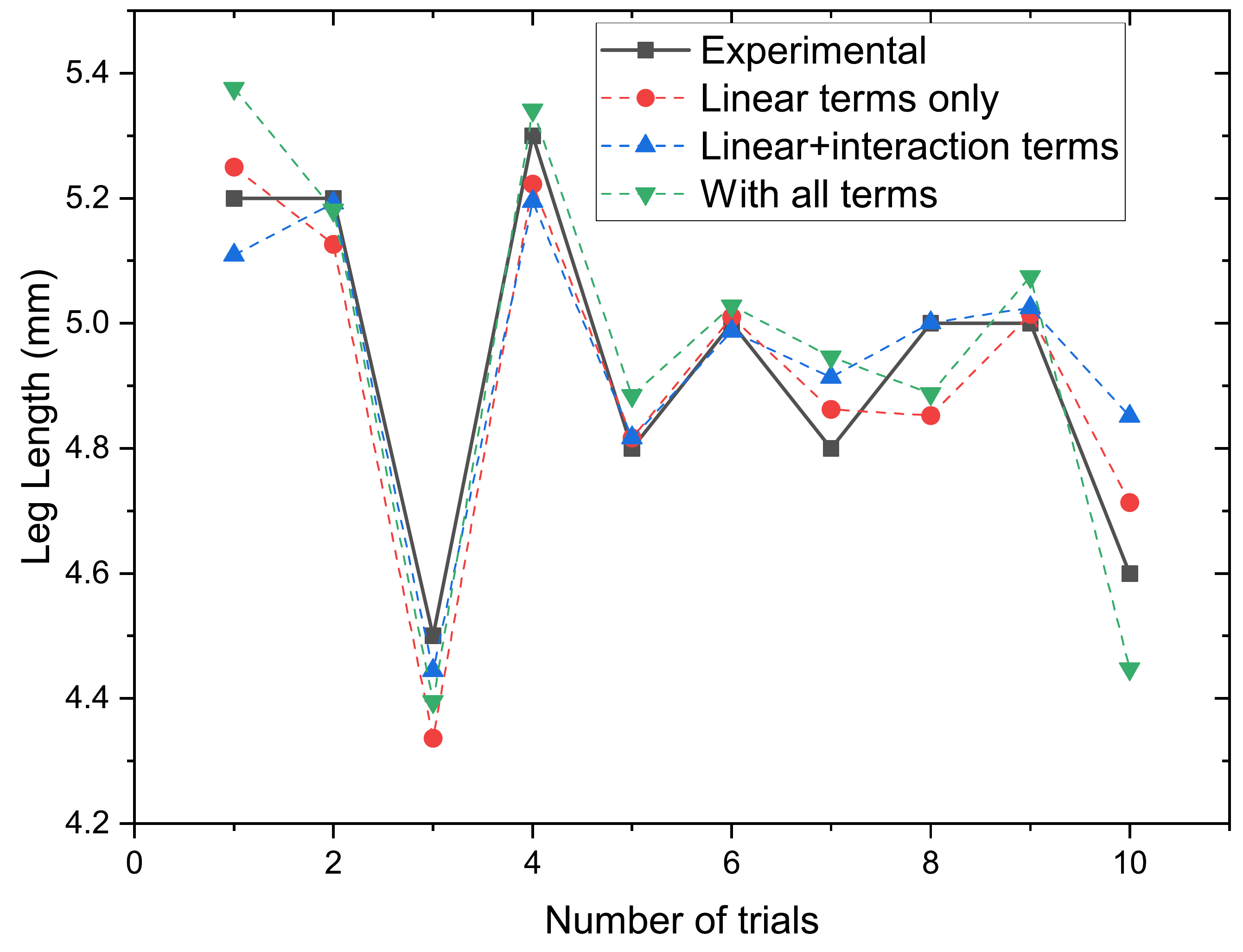}
		\label{fig:picture12a}}
	\subfloat[]{
		\includegraphics[width=0.3\linewidth, height=0.2\textwidth]{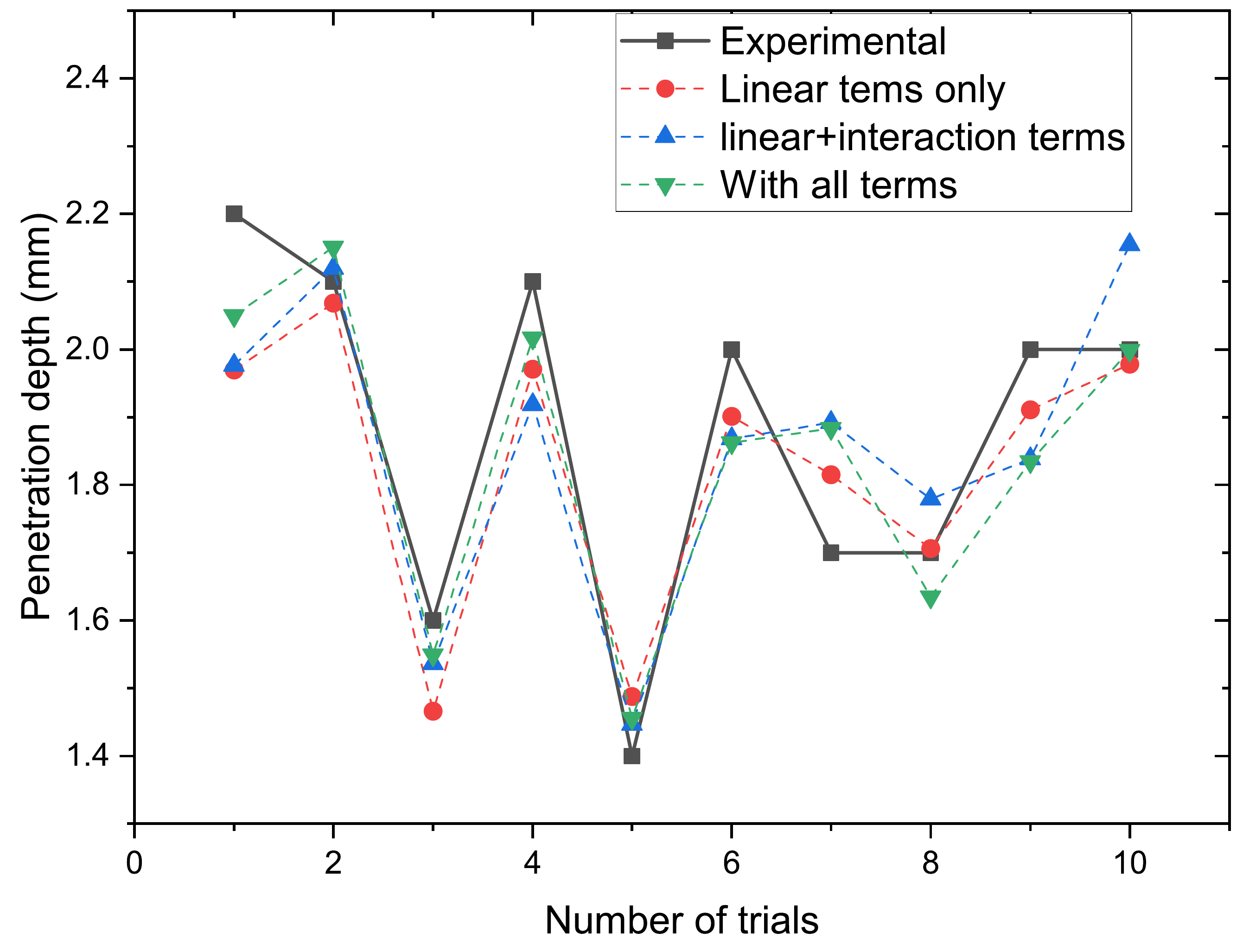}
		\label{fig:picture12b}}
	
	\subfloat[]{
		\includegraphics[width=0.3\linewidth, height=0.2\textwidth]{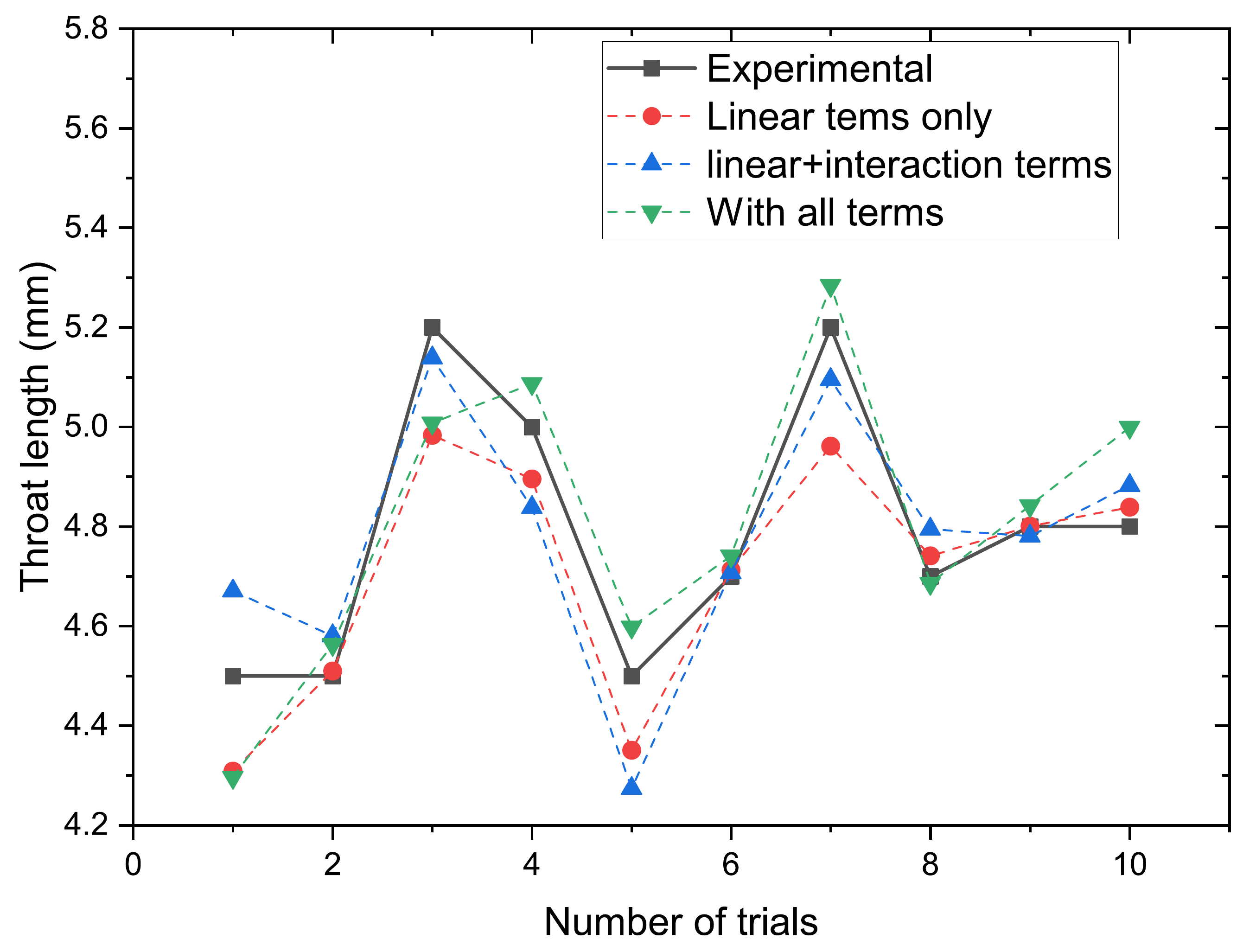}
		\label{fig:picture12c}}
	\subfloat[]{
		\includegraphics[width=0.3\linewidth, height=0.2\textwidth]{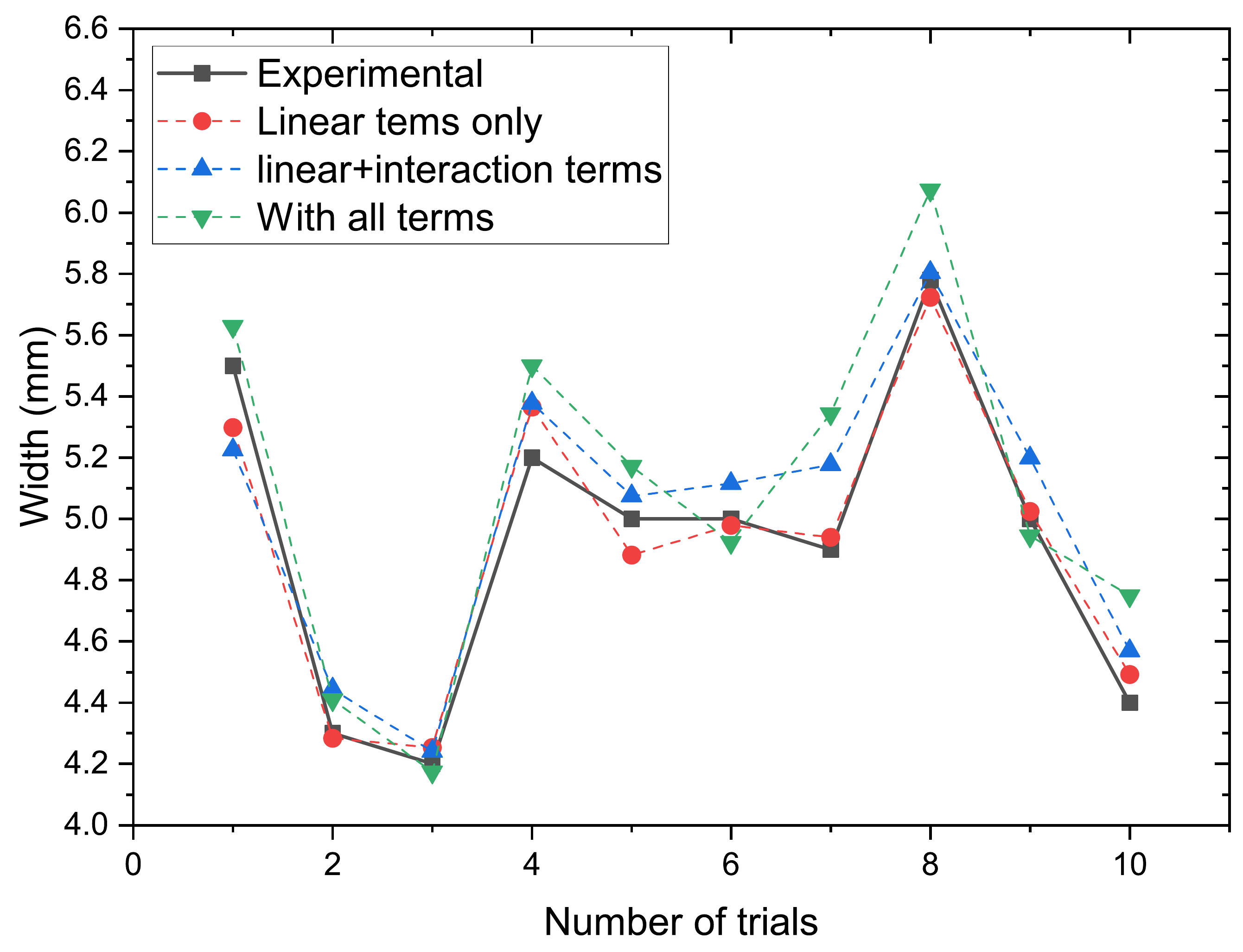}
		\label{fig:picture12d}}
	\caption{Experimental vs predicted values for ANN model }
	\label{fig:picture12}
\end{figure*}
\begin{figure*}[tph]
	\centering
	\subfloat[]{%
		\includegraphics[width=0.3\linewidth, height=0.2\textwidth]{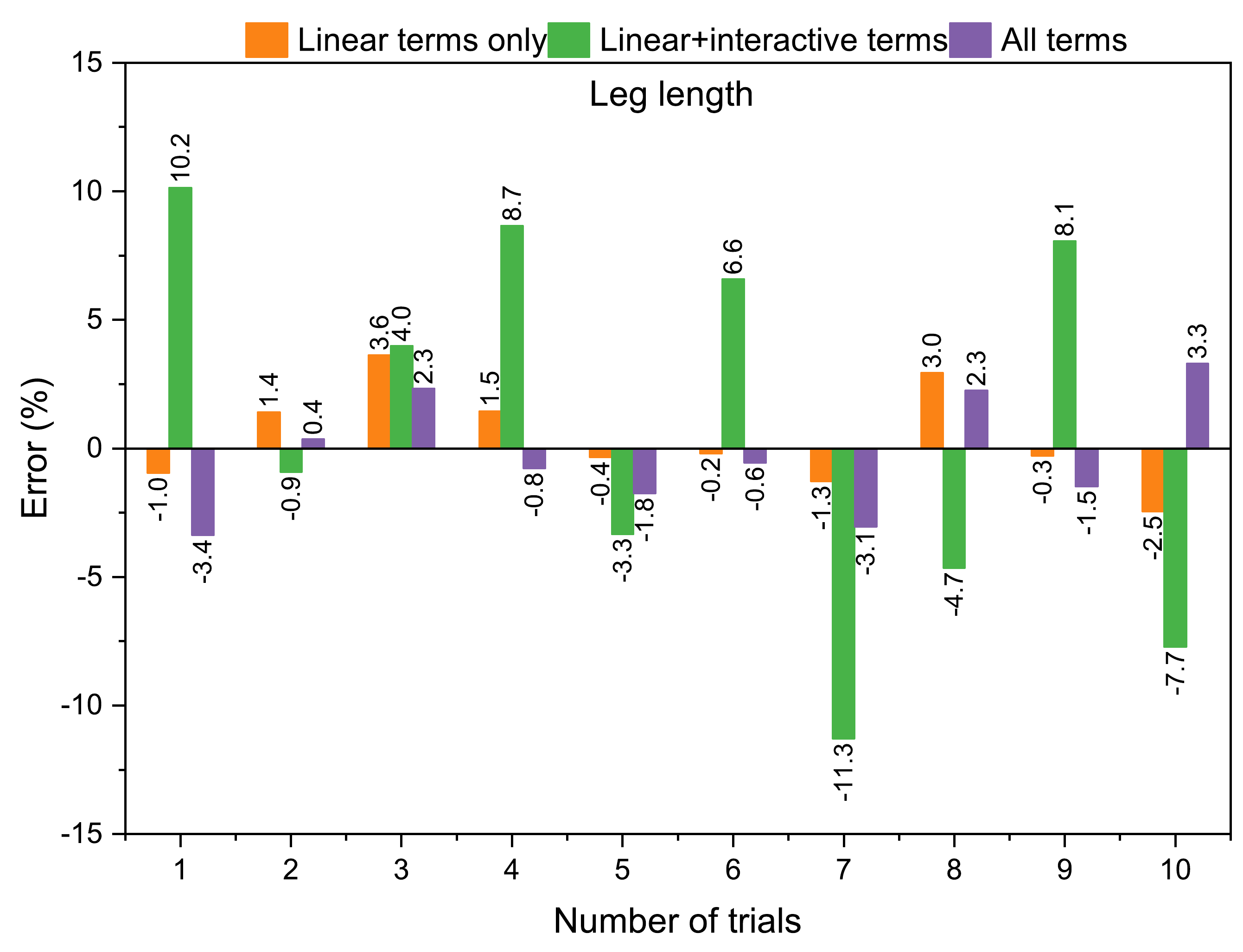}
		\label{fig:picture13a}}
	\subfloat[]{
		\includegraphics[width=0.3\linewidth, height=0.2\textwidth]{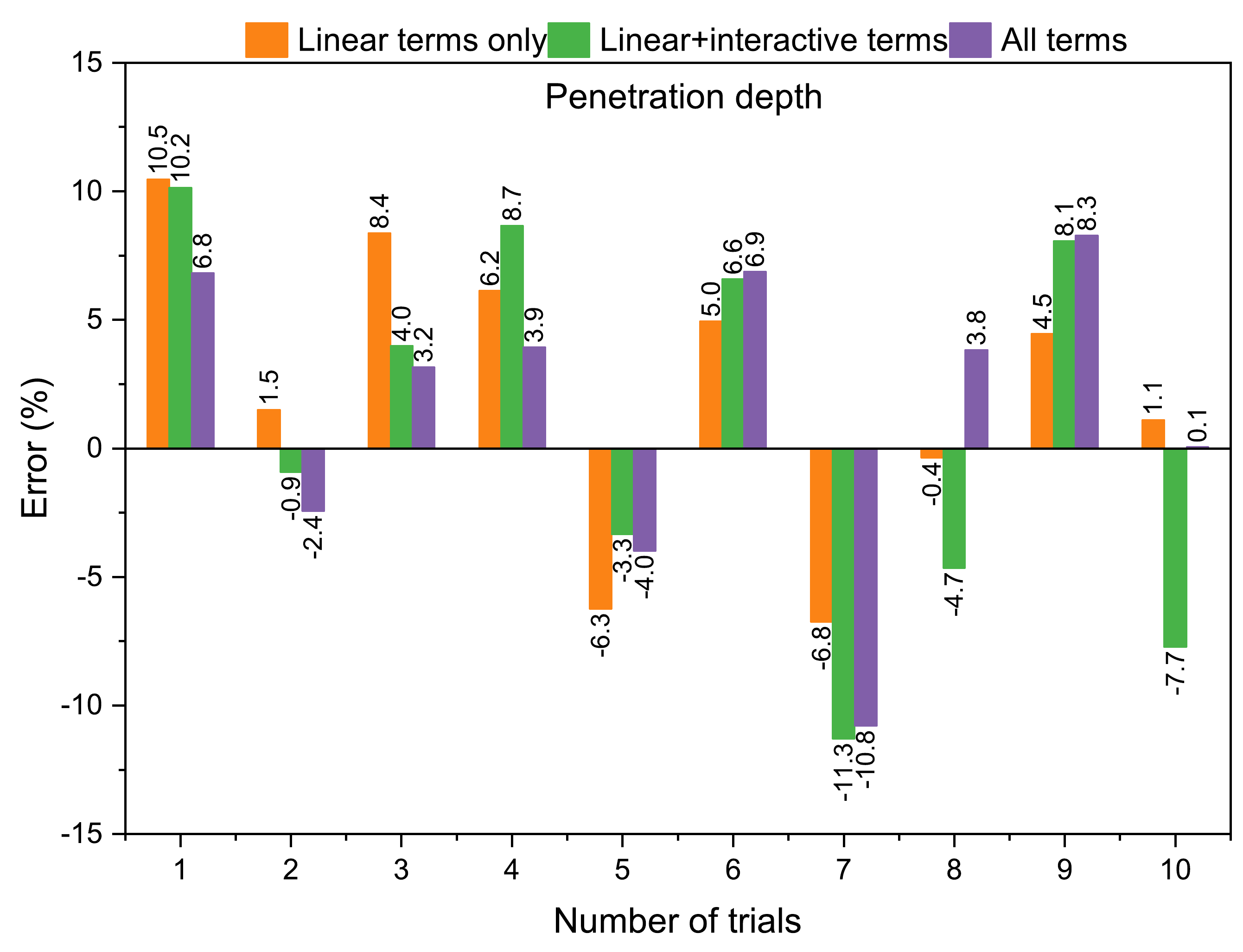}
		\label{fig:picture13b}}
	
	\subfloat[]{
		\includegraphics[width=0.3\linewidth, height=0.2\textwidth]{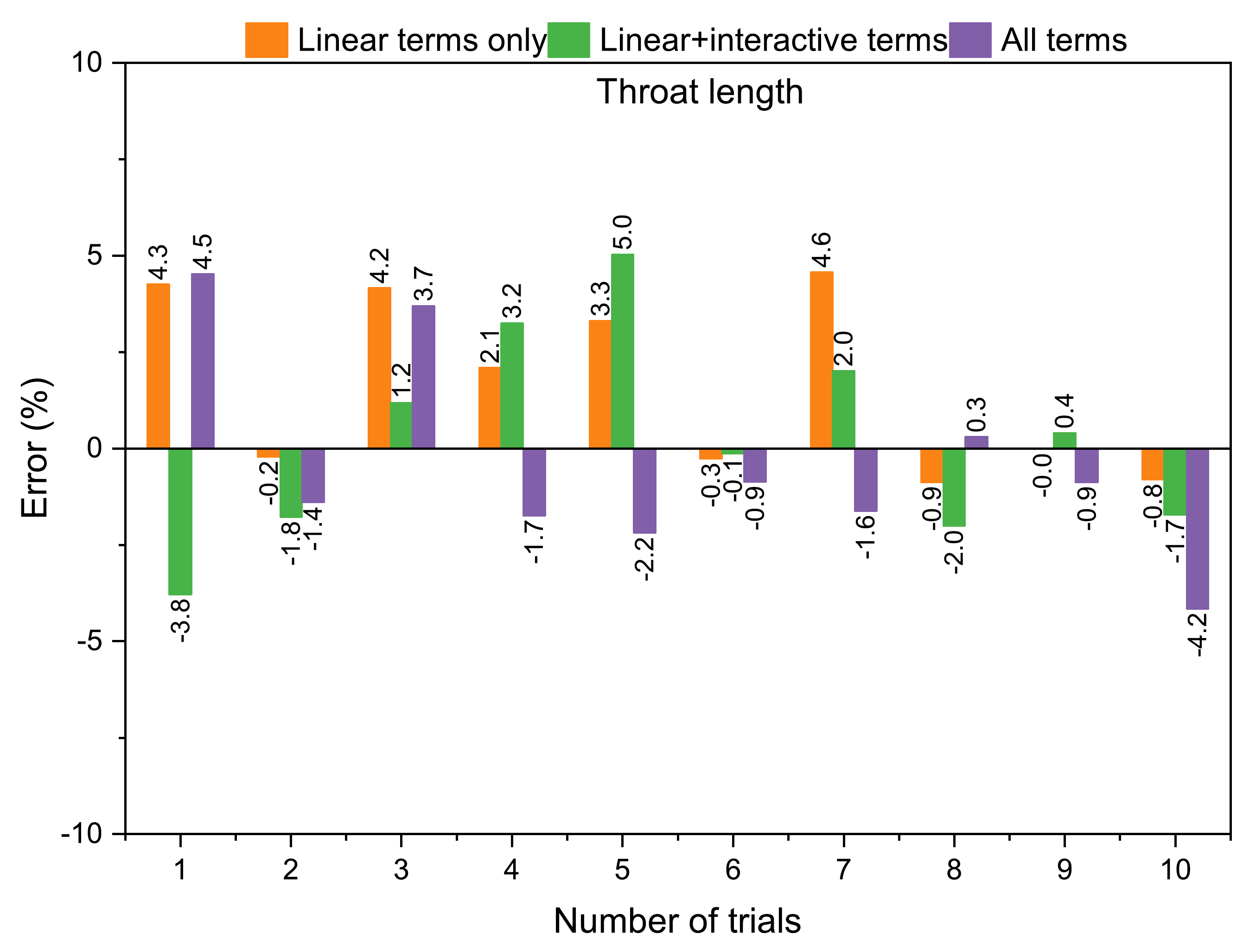}
		\label{fig:picture13c}}
	\subfloat[]{
		\includegraphics[width=0.3\linewidth, height=0.2\textwidth]{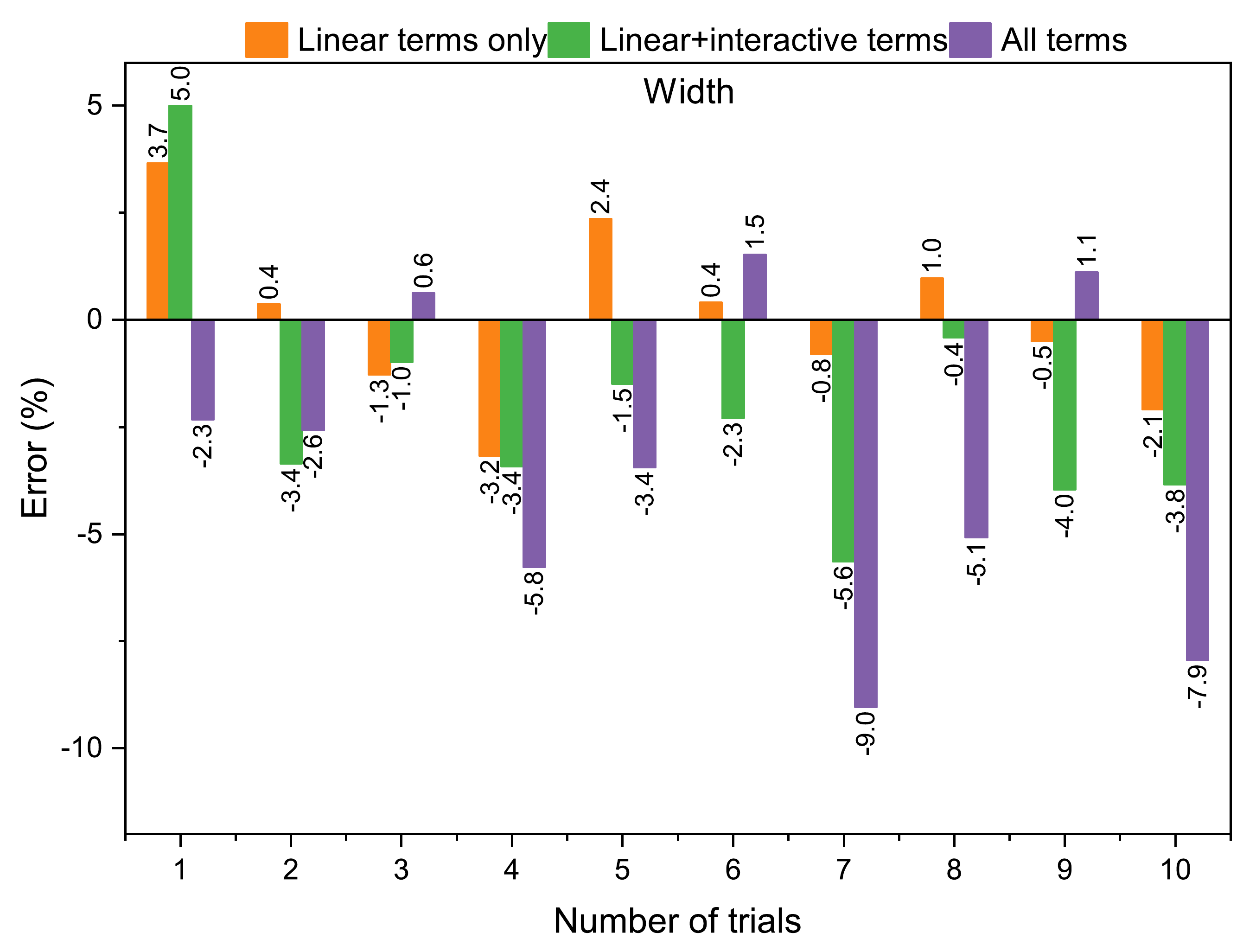}
		\label{fig:picture13d}}
	\caption{Percentage of error in ANN prediction }
	\label{fig:picture13}
\end{figure*}
\begin{figure*}[tph]
	\centering
	\subfloat[]{%
		\includegraphics[width=0.4\linewidth, height=0.3\textwidth]{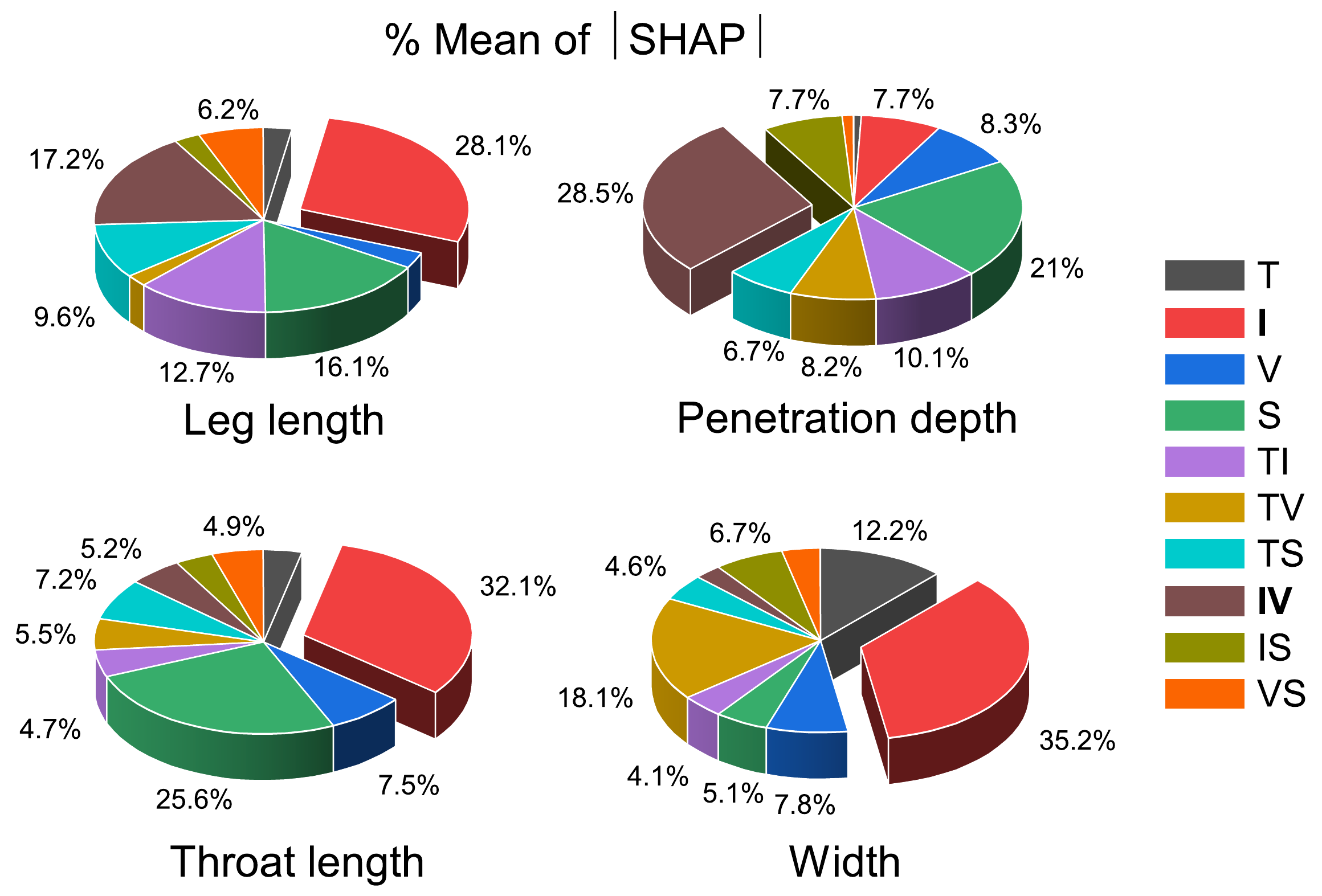}
		\label{fig:picture14a}}
	\subfloat[]{
		\includegraphics[width=0.4\linewidth, height=0.3\textwidth]{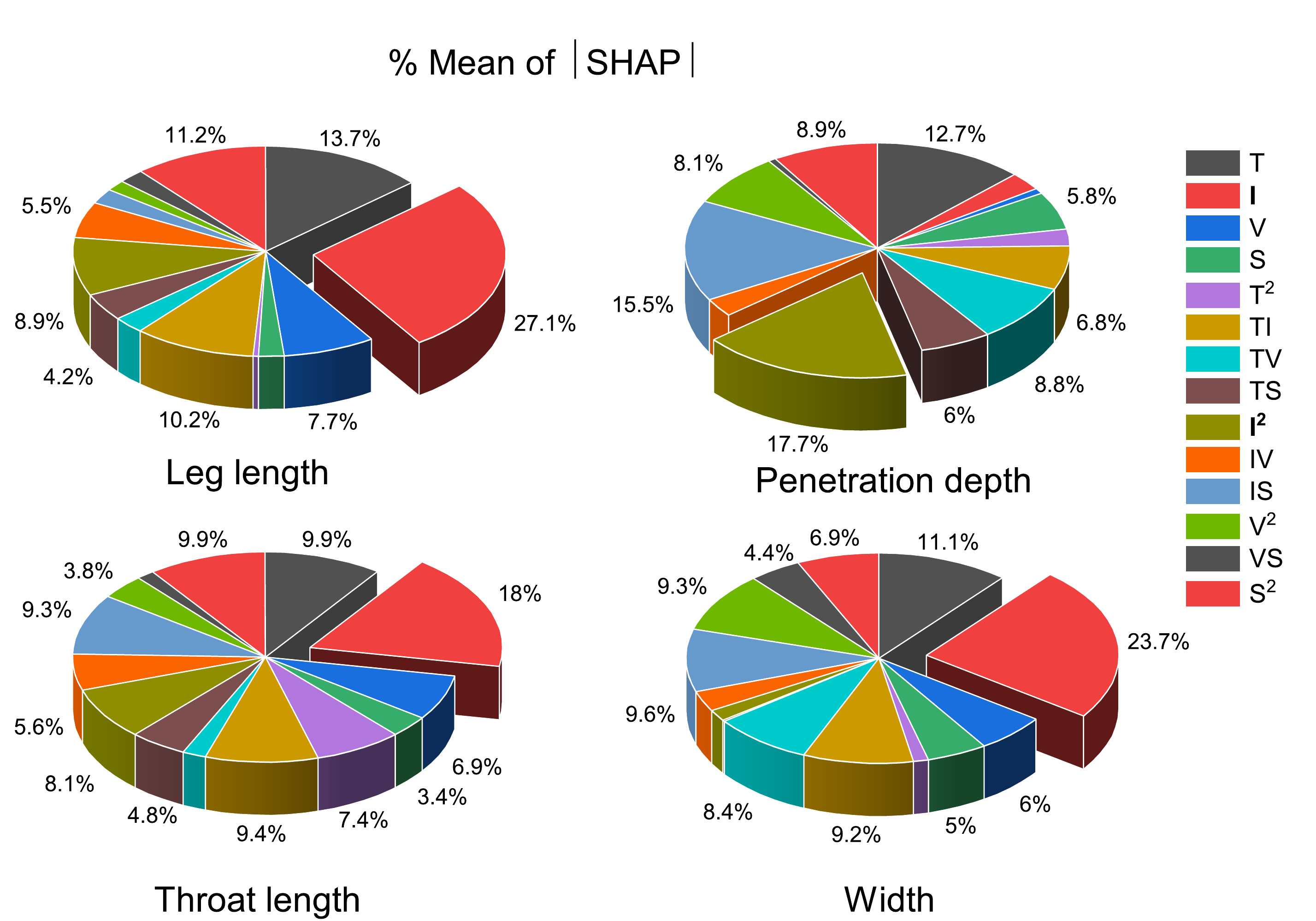}
		\label{fig:picture14b}}
	\caption{Feature importance on ANN model(a) with linear and interactive terms (b) with all terms }
	\label{fig:picture14}
\end{figure*}

Figure \ref{fig:picture10} shows the \% of error in predicted outputs performed on test datasets. In each graph, \% of error obtained from three sets of regression relations are plotted. For leg length and depth of penetration, the error percentage increased as the interactive terms are introduced (Figure \ref{fig:picture10a} and \ref{fig:picture10b}). But in contrast, for throat length and width, the error percentage decreased significantly as the higher order terms are included (Figure \ref{fig:picture10c} and \ref{fig:picture10d}). Maximum percentage of error (14.2\%) is observed in case of MLR for width prediction. In most of the cases error percentage is below 10\% which can be regarded as satisfactory in nature. 

The SHAP plots shown in Figure \ref{fig:picture11a} and \ref{fig:picture11b} explains the impact of different input features on the prediction made by the second and the third sets of MLR models. It shows the pie-chart of the average of modulus of SHAP value for output parameters against input parameters.
 
In figure \ref{fig:picture11a} , the plot explains that the circuit current has a high impact on the prediction of all bead shape parameters. Welding speed is highly influential for determining the throat and the leg length. Plate thickness has least impact on all the output parameters. Figure \ref{fig:picture11b} shows the similar trend in the feature importance when all the terms are considered. In this case, leg length is mostly influenced by combined effect of voltage and welding speed. Width and depth of penetration are impacted by circuit current. Throat length is highly affected by welding speed.

Figures \ref{fig:picture12a}-\ref{fig:picture12d} shows the comparison between experimental and predicted value (obtained from all the three ANN models) for the trial datasets provided on the Table \ref{table2}. For leg length and penetration depth, a slight difference in the predicted results is observed while considering interactive terms.The \% of error is also high as compared to other ANN models (Figure \ref{fig:picture13a},\ref{fig:picture13b}). But in other two parameters shows the deviation with in the range of 10\% (Figure \ref{fig:picture13c},\ref{fig:picture13d}) .The overall training record of the train datasets is shown in performance plot (Figure \ref{fig:picture8}),which clearly shows the error reduces after more epochs of training.

In figure \ref{fig:picture14a}, the plot shows that circuit current is highly influential for all output prediction in ANN model with interactive terms. For the depth of penetration, it is the combination of circuit current and voltage which affects the most. Again plate thickness has least impact on all the output parameters as in MLR models. Figure \ref{fig:picture14b} shows the similar trend in the feature importance when all the terms are considered in ANN model. The above results are informative and provide much-needed information on the impactful features of the outcome of the model. This corroborates with the fact that current (Amp) is the major factor that influences the welding process since the total heat input is directly proportional to the square of the current (Joule heating) followed by voltage and welding speed.

\section{Conclusion}
Based on the different welding input parameter ranges (plate thickness 3-10 mm, welding current 125-310 Amp, arc voltage 18-36 V and speed of weld 3.25-9.25 mm/s), three sets of multiple linear regression and artificial neural network models are developed and used to predict the bead geometry parameters for GMAW welding process. The following conclusions are drawn by comparing and analyzing the models:
\begin{itemize}
	\item Both multiple linear regression and ANN model has a good correlation between the welding process parameters and the bead geometry considering the linear terms only. 
	\item Statistically ANN model is more predictable than MLR model when only the linear terms are considered.
	\item The performance of the MLR model enhanced as the higher order terms are included. At the same time performance of the ANN model is unaffected.
	\item Hence proposed MLR models can be more effective as compared to ANN models when higher ordered terms are considered and effectively used as a prediction tool for bead geometry parameters for GMAW for being used in a numerical model.
	\item The outcomes from the SHAP values indicated that circuit current is impactful on the bead shape parameters followed by arc voltage, welding speed and plate thickness. 
	\item This shows that the predictive models can also provide valuable insight on the combined effect of several input variables on the weld bead quality.
	\item The overall performance of the predictive models show that these can be used for determining the weld bead size to be incorporated in a simulation for more accurate prediction of weld induced distortion and stresses. 
	
\end{itemize}
\section*{Conflict of interest}
The authors declare that there is no conflict
of interest regarding the publication of the papers.

\bibliography{sn-bibliography}


\end{document}